\documentclass{article}
\usepackage{ijcai26}
\usepackage{times}
\usepackage{soul}
\usepackage{url}
\usepackage[hidelinks]{hyperref}
\usepackage[utf8]{inputenc}
\usepackage[small]{caption}
\usepackage{graphicx}
\usepackage{amsmath}
\usepackage{amssymb}
\usepackage{amsthm}
\usepackage{booktabs}
\usepackage{algorithm}
\usepackage{algorithmic}
\usepackage{xcolor}
\usepackage{tikz}
\usetikzlibrary{arrows.meta,positioning,calc,fit,backgrounds}
\usepackage[switch]{lineno}

\newtheorem{theorem}{Theorem}
\newtheorem{proposition}{Proposition}
\newtheorem{corollary}{Corollary}
\newtheorem{lemma}{Lemma}
\newtheorem{assumption}{Assumption}

\title{When the Proxy Is Wrong:\\Mismatch-Corrected Bandits for Adaptive Data Sampling}

\author{
Amine Saadi$^{1,2}$
\and
Elissa Mhanna$^1$\and
Raphaël Féraud$^1$\And
Gwénolé Lecorvé$^1$\\
\affiliations
$^1$Orange Research, Lannion, France\\
$^2$École Polytechnique, Institut Polytechnique de Paris, Palaiseau, France\\
\emails
amine.saadi.2023@polytechnique.org, \{elissa.mhanna, raphael.feraud, gwenole.lecorve\}@orange.com
}

\begin{document}

\maketitle

\begin{abstract}
Data allocation methods often use a small proxy model to select training data for a larger target model. However, this proxy can rank data differently from the target, leading to suboptimal training performance. To address this problem, we model the mismatch between the proxy and target models and incorporate it into the allocation decision. To do so, the problem is cast as a multi-fidelity bandit, where each arm is a data intervention. The proxy reward is observed every round, whereas the target reward is revealed only on rare audits of the same update. These paired observations identify arm-wise mismatches without assuming a shared offset or preservation of rankings.
Mismatch-corrected bandits correct the proxy with paired residuals, audit only uncertain decisions, and pace the audit budget over training. From a theoretical viewpoint, we prove a regret decomposition and lower bounds under stationarity. These bounds show that proxy-only allocation can incur linear regret and that a minimum number of paired audits is necessary. Under abrupt changes, using the audit budget early can also incur linear regret. A sliding-window policy that spreads audits over training achieves regret of order $\sqrt{\Gamma_T T\log T}$ where $\Gamma_T$ denotes the number of change points. Experimentally, we test the approach in a closed-loop proof of concept on CivilComments-WILDS, where BERT-Small guides the training of BERT-Large. Mismatch correction improves worst-group accuracy over a target-only policy with the same target-model budget. Spreading audits over training improves worst-group accuracy by 6.3 to 10.5 percentage points at the same audit count.
\end{abstract}

%======================================================================
\section{Introduction}
\label{sec:intro}

Language models are trained on heterogeneous mixtures of web text, code, books, dialogue, and domain- or language-specific collections, and the way the optimization budget is distributed across these sources shapes the capabilities of the final model. Because the effect of a data choice on a large model is expensive to measure, most allocation methods rely on a cheaper signal: a small proxy model, a reference model, or training-time statistics~\cite{xie2023doremi,liu2025regmix,albalak2023odm,chen2025aioli}. The implicit assumption is that this signal ranks the candidate data as the target model would.

This assumption can fail even when the proxy is informative. A proxy may be accurate on average and still attribute the largest gains to the wrong data, because its error depends on the data group. In our proof of concept, the rewards of a BERT-Small proxy and of a BERT-Large target are positively correlated, yet about one third of the per-round gains have opposite signs for the two models. A policy that trusts the proxy can then estimate its own signal very precisely while repeatedly favoring data helping the proxy rather than the target.

\begin{figure*}[t]
    \centering
    \resizebox{\textwidth}{!}{%
\begin{tikzpicture}[
    font=\scriptsize, >={Stealth[length=3.5pt]},
    panel/.style={draw=black!40, rounded corners=3pt, minimum width=3.25cm, minimum height=3.05cm, fill=black!2},
    ptitle/.style={font=\scriptsize\bfseries, anchor=north, text=black!85},
    tile/.style={draw=#1!70!black, fill=#1!18, rounded corners=1pt, minimum width=5.5mm, minimum height=4mm, inner sep=0pt, font=\scriptsize},
    circ/.style={circle, draw=#1!70!black, fill=#1!18, minimum size=4.6mm, inner sep=0pt, font=\scriptsize},
    model/.style={draw=#1, line width=0.8pt, rounded corners=2pt, fill=white, minimum width=12mm, minimum height=6mm, align=center},
    note/.style={align=center, text=black!70, font=\tiny}]
  \definecolor{smallc}{RGB}{42,120,214}
  \definecolor{largec}{RGB}{235,104,52}
  \colorlet{ga}{smallc} \colorlet{gb}{largec} \definecolor{gc}{RGB}{27,175,122} \definecolor{gk}{RGB}{140,110,190}
  % ---- panels
  \foreach \i/\x in {1/0,2/3.6,3/7.2,4/10.8,5/14.4}{\node[panel] (P\i) at (\x,0) {};}
  \node[ptitle] at ($(P1.north)+(0,-0.08)$) {1. Candidate data groups};
  \node[ptitle] at ($(P2.north)+(0,-0.08)$) {2. Train both models};
  \node[ptitle] at ($(P3.north)+(0,-0.08)$) {3. Observe the gains};
  \node[ptitle] at ($(P4.north)+(0,-0.08)$) {4. Correct the ranking};
  \node[ptitle] at ($(P5.north)+(0,-0.08)$) {5. Choose the next group};
  % ---- 1. groups
  \node[tile=ga] at (-0.55,0.45) {$G_1$}; \node[tile=gb] at (0.55,0.45) {$G_2$};
  \node[tile=gc] at (-0.55,-0.25) {$G_3$}; \node at (0,-0.25) {$\cdots$}; \node[tile=gk] at (0.55,-0.25) {$G_K$};
  \node[note] at (0,-0.95) {overlapping groups,\\e.g.\ domains or subpopulations};
  % ---- 2. train
  \node[tile=gb, minimum width=8mm] (blkA) at (3.2,0.55) {$G_2$};
  \node[draw=black!50, fill=black!8, rounded corners=1pt, minimum width=8mm, minimum height=4mm, right=0pt of blkA, font=\tiny] (blkB) {pool};
  \coordinate (blk) at (blkA.south east);
  \node[model=smallc] (sm) at (2.85,-0.45) {Small\\model};
  \node[model=largec] (lg) at (4.35,-0.45) {Large\\model};
  \draw[->] (blk) -- ++(0,-0.15) -| (sm.north);
  \draw[->] (blk) -- ++(0,-0.15) -| (lg.north);
  \node[note] at (3.6,-1.15) {half chosen group, half random pool data\\same block for both models};
  % ---- 3. gains
  \draw[->, smallc, line width=0.8pt] (6.05,0.55) -- (6.75,0.55);
  \node[anchor=west, align=left, font=\tiny] at (6.8,0.55) {small-model gain $Y_t$\\\textbf{every round}};
  \draw[->, largec, line width=0.8pt, dashed] (6.05,-0.2) -- (6.75,-0.2);
  \node[anchor=west, align=left, font=\tiny] at (6.8,-0.2) {large-model gain $Z_t$\\\textbf{only if audited}};
  \node[draw=black!50, rounded corners=1.5pt, fill=white, align=center, font=\tiny, text width=2.8cm] at (7.2,-1.0)
       {audit only uncertain decisions,\\spreading the budget over training};
  % ---- 4. ranking correction
  \node[note, anchor=east] at (9.95,0.6) {small:};
  \node[circ=ga] at (10.2,0.6) {$1$}; \node at (10.6,0.6) {$>$};
  \node[circ=gb] at (11.0,0.6) {$2$}; \node at (11.4,0.6) {$>$};
  \node[circ=gc] at (11.8,0.6) {$3$};
  \node[note, anchor=east] at (9.95,0.0) {large:};
  \node[circ=gb] at (10.2,0.0) {$2$}; \node at (10.6,0.0) {$>$};
  \node[circ=gc] at (11.0,0.0) {$3$}; \node at (11.4,0.0) {$>$};
  \node[circ=ga] at (11.8,0.0) {$1$};
  \node[draw=black!50, rounded corners=1.5pt, fill=white, align=center, font=\tiny, text width=2.9cm] at (10.8,-0.85)
       {estimate the gap $Z_t-Y_t$ per group\\on audited rounds, add it to the\\small-model estimate};
  % ---- 5. choice
  \node[draw=black!50, rounded corners=1.5pt, fill=white, align=center, font=\tiny, text width=2.8cm] at (14.4,0.35)
       {optimistic score per group,\\built from the corrected estimate\\and the direct audits};
  \node[tile=gb, minimum width=8mm] (nxA) at (14.0,-0.6) {$G_2$};
  \node[draw=black!50, fill=black!8, rounded corners=1pt, minimum width=8mm, minimum height=4mm, right=0pt of nxA, font=\tiny] {pool};
  \node[note] at (14.4,-1.05) {next block};
  % ---- flow
  \foreach \a/\b in {1/2,2/3,3/4,4/5}{\draw[->, black!60, line width=0.6pt] (P\a.east) -- (P\b.west);}
  \draw[->, black!60, line width=0.6pt, dashed] (P5.south) -- ++(0,-0.3) -| node[note, pos=0.25, below] {next round} (P2.south);
\end{tikzpicture}%
    }
    \caption{Overview of mismatch-corrected adaptive data sampling. At each round, a controller chooses a data group. The next training block mixes examples from this group with random examples from the whole pool and updates both a small proxy model and the large target model (1--2). The proxy's gain $Y_t$ is observed every round, whereas the target's gain $Z_t$ is revealed only on rare audits, requested when a decision is still uncertain and spread over training (3). The two models may rank the groups differently. On audited rounds, the controller measures the gap $Z_t-Y_t$ for the selected group and uses it to correct the proxy ranking (4). The controller then chooses the next group from an optimistic score built on the corrected estimate (5).}
    \label{fig:teaser}
\end{figure*}
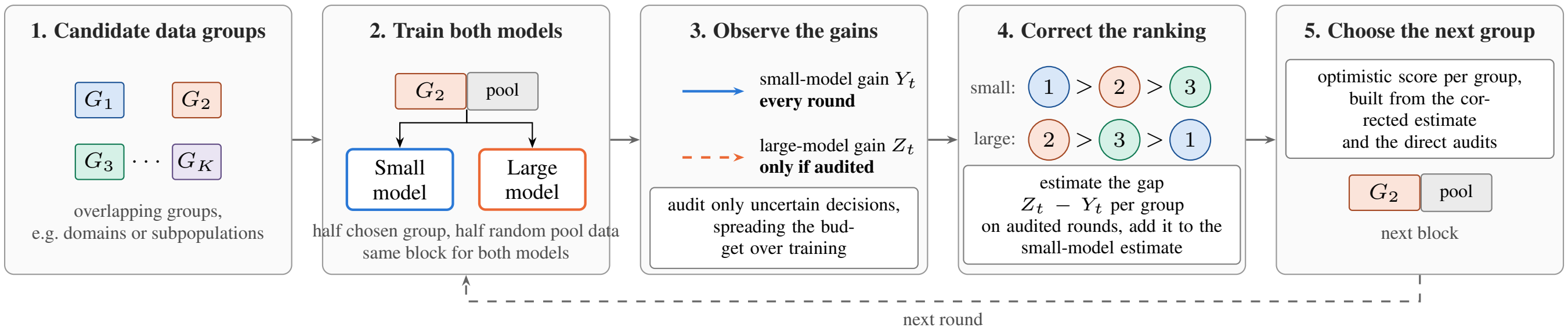

We study this problem as a multi-fidelity bandit (Figure~\ref{fig:teaser}). Each arm represents a data intervention. After every update, the controller observes the proxy reward. On rare audited updates, it also observes the target reward for the same intervention and update. In our proof of concept, one audit means evaluating BERT-Large (335M parameters) on 20{,}005 validation examples, against BERT-Small (28.8M) on 2{,}001 examples for the proxy. Pairing the two observations makes the proxy error identifiable arm by arm: instead of replacing the proxy with sparse target feedback, the audits estimate the residual between the two signals and correct the ranking induced by the proxy. Multi-fidelity bandits assume that the bias of each fidelity is bounded by a known constant~\cite{kandasamy2016multifidelity}, and bandits with sparse and noisy feedback assume an offset common to all arms~\cite{bouneffouf2025sparse}. Our setting makes neither assumption: the proxy--target mismatch may vary across arms and can therefore reverse their ranking.

A second question then arises: \emph{when} should audits be spent? If the relation between proxy and target were fixed, the answer would be as early as possible, since an early audit is then as informative as a late one. During training, however, this relation drifts with the state of the models. A calibration obtained early can become stale, and a controller that consumes its budget on the first uncertain decisions cannot revise them later. At identical audit count, the timing of target feedback can matter as much as its amount.

\paragraph{Contributions.}
\begin{itemize}
    \item We formulate adaptive data sampling with a cheap proxy and rare target evaluations as a multi-fidelity bandit with an arm-dependent, possibly rank-inverting mismatch and paired audits (Section~\ref{sec:problem}).
    \item We propose mismatch-corrected bandits that combine a corrected confidence index, a query gate that requests target feedback only when the mismatch uncertainty can change the decision, and audit pacing under a hard budget (Section~\ref{sec:method}).
    \item We analyze these policies (Section~\ref{sec:analysis}). Under stationarity, a regret decomposition isolates the cost of deferred audits, and lower bounds show that proxy-only allocation incurs linear regret and that $\Omega(\sigma_\beta^2/\Delta^2)$ paired audits are necessary. Under piecewise stationarity with $\Gamma_T$ changes, we show that front-loaded audits incur linear regret, and that pacing combined with a sliding window achieves $O(\sqrt{\Gamma_TT\log T})$ regret.
    \item We map the regimes in which correction pays off on synthetic environments, and validate the mechanism in a closed loop where both models are actually trained: at equal target budget, correction improves worst-group accuracy over a target-only policy by $+2.5$ points, and at identical audit count, pacing improves it by $+6.3$ to $+10.5$ points (Section~\ref{sec:experiments}).
\end{itemize}

%======================================================================
\section{Related Work}
\label{sec:related}
 
\paragraph{Adaptive data allocation.}
Choosing how much optimization budget each part of a training corpus receives has been studied under several names: data mixture optimization, curriculum learning, data selection, and online reweighting~\cite{saadi2026survey}. Offline mixture methods estimate a fixed allocation before the main run, often from a small proxy model: DoReMi uses excess loss relative to a reference model~\cite{xie2023doremi}, DoGE uses gradient alignment with a generalization target~\cite{fan2024doge}, and RegMix or data mixing laws fit a predictor from proxy runs~\cite{liu2025regmix,ye2025mixinglaws}. Online methods revise the allocation during training: automated curricula and ODM cast the choice of task or domain as a bandit problem~\cite{graves2017automated,albalak2023odm}, while Aioli and ADO estimate the marginal effect of each domain from training dynamics~\cite{chen2025aioli,jiang2025ado}. Across these families, the utility signal is a proxy for the effect on the model of interest, and the residual error between this proxy and a higher-fidelity measurement is rarely modeled explicitly. Our work targets precisely this residual when it changes the allocation decision.
 
\paragraph{Multi-fidelity and auxiliary-feedback bandits.}
Multi-fidelity bandits combine cheap approximations with costly exact evaluations, typically assuming that the bias of each fidelity is bounded by a known constant and letting the learner choose the fidelity~\cite{kandasamy2016multifidelity}. Bouneffouf et al.~\shortcite{bouneffouf2025sparse} study a frequent noisy reward complemented by a sparse reliable one. They assume that the two means differ by the same offset for every arm, so both signals rank the arms in the same order. We allow the proxy--target mismatch to differ across arms, which can reverse their ranking. We estimate this mismatch from paired observations of both signals on the same pull. Our corrected estimator has the same form as the rectifier of prediction-powered inference~\cite{angelopoulos2023ppi} and its active variants~\cite{zrnic2024active}. We use this estimator inside a sequential decision rule that also determines when to acquire costly target feedback. Variance-adaptive confidence bounds follow UCB-V~\cite{audibert2009ucbv}.
 
\paragraph{Non-stationary bandits and budget pacing.}
When rewards change over time, discounted and sliding-window UCB forget old observations to track the current best arm \cite{garivier2011switching}. Our setting adds a second resource, the audit budget, whose consumption over time matters as much as its total: pacing rules that spread a hard budget over the remaining horizon are standard in online advertising~\cite{balseiro2019learning}, and budgeted feedback relates to label-efficient prediction~\cite{cesabianchi2005label} and bandits with knapsacks~\cite{badanidiyuru2018knapsacks}. We combine an uncertainty-driven query rule with a simple pacing rule, quantify the regret incurred when pacing defers a query, and extend the sliding-window analysis of~\cite{garivier2011switching} to a setting where the second feedback channel is budgeted.

%======================================================================
\section{Problem Formulation}
\label{sec:problem}
 
\paragraph{Interventions.}
Let $\mathcal{D}$ be a pool of candidate training data and $[K]=\{1,\dots,K\}$ a finite library of \emph{interventions}. Intervention $k$ is a sampling distribution $w_k$ over $\mathcal{D}$, for instance one that over-represents a domain, a language, or a subpopulation. The underlying groups need not be disjoint. At each round $t=1,\dots,T$, a controller selects $A_t\in[K]$, a training block $\mathcal{B}_t\sim w_{A_t}$ is drawn, and this \emph{same} block is used to update both a small proxy model and the large target model. The action is therefore a decision about the distribution of the next updates rather than the selection of a single example.
 
\paragraph{Two-fidelity feedback.}
Every round reveals a proxy reward $Y_t$, the improvement of the proxy model on a proxy validation set after the update. The target reward $Z_t$, the improvement of the target model on a target validation set after the same update, is expensive and is revealed only when the controller \emph{audits} the round. We denote by $\mathcal{H}_{t-1}$ the history up to round $t-1$. Intervention $k$ has proxy mean $p_k$ and target mean $r_k$,
\begin{align}
    p_k&=\mathbb{E}[Y_t\mid\mathcal{H}_{t-1},A_t=k],\\
    r_k&=\mathbb{E}[Z_t\mid\mathcal{H}_{t-1},A_t=k],
\end{align}
where $Z_t$ denotes the target feedback that the update would produce whether or not it is audited. The \emph{mismatch} of intervention $k$ is
\begin{equation}
    \beta_k = r_k - p_k .
\end{equation}
We make no assumption on the sign of $\beta_k$, on a common offset across arms, or on rank preservation: the proxy-best and target-best interventions may differ.
 
\paragraph{Paired audits and budget.}
Before observing the current feedback, the controller chooses an audit indicator $Q_t\in\{0,1\}$ under the hard budget $\sum_{t=1}^{T}Q_t\le B$. When $Q_t=1$, both rewards are observed for the same intervention and the same update, which yields the paired difference
\begin{equation}
    D_t = Z_t - Y_t = \beta_{A_t} + \eta_t ,
\end{equation}
with $\eta_t$ a centered noise. Pairing is what makes the mismatch identifiable arm by arm. The budget $B$ limits the amount of target \emph{information} revealed to the controller, not the number of target updates: the target model is trained on every block.
 
\begin{assumption}[Stationary paired feedback]
\label{ass:model}
The means $p_k$ and $r_k$ do not depend on $t$ or on the history. Conditionally on $\mathcal{H}_{t-1}$ and on $A_t=k$, the proxy noise $Y_t-p_k$, the paired residual noise $\eta_t$, and the target noise $Z_t-r_k$ are centered and sub-Gaussian with parameters $\sigma_p$, $\sigma_\beta$, and $\sigma_r$. The pair $(A_t,Q_t)$ is chosen from $\mathcal{H}_{t-1}$ and independent internal randomization before the current feedback is observed.
\end{assumption}
Section~\ref{sec:analysis-piecewise} relaxes the first condition to means that change at unknown rounds.
 
\paragraph{Objective.}
Let $k^\star=\arg\max_k r_k$ be the target-optimal intervention, assumed unique, and $\Delta_k=r_{k^\star}-r_k$. The controller minimizes the target pseudo-regret
\begin{equation}
    R_T=\sum_{t=1}^{T}\big(r_{k^\star}-r_{A_t}\big)=\sum_{k\neq k^\star}\Delta_k N_T(k),
    \label{eq:regret}
\end{equation}
where $N_T(k)$ is the number of rounds on which intervention $k$ is played. A policy that optimizes the proxy alone may estimate $p_k$ accurately and still play a target-suboptimal intervention almost every round (Section~\ref{sec:analysis}).

%======================================================================
\section{Mismatch-Corrected Bandits}
\label{sec:method}
 
\subsection{Mismatch-Corrected Index}
\label{sec:method-index}
 
The method keeps three statistics per intervention. With $N_{t-1}(k)$ pulls (rounds on which $k$ was played) and $M_{t-1}(k)$ audits of intervention $k$ before round $t$,
\begin{align}
    \hat p_t(k)&=\frac{1}{N_{t-1}(k)}\sum_{s<t,\,A_s=k}Y_s,\\
    \hat\beta_t(k)&=\frac{1}{M_{t-1}(k)}\sum_{s<t,\,A_s=k,\,Q_s=1}(Z_s-Y_s),\\
    \hat r_t(k)&=\frac{1}{M_{t-1}(k)}\sum_{s<t,\,A_s=k,\,Q_s=1}Z_s .
\end{align}
The decomposition $r_k=p_k+\beta_k$ suggests estimating the target value by $\hat p_t(k)+\hat\beta_t(k)$: the proxy term uses every pull and quickly separates the interventions, while the rare audits are spent only on the part of the target utility that the proxy does not capture. Writing $\bar Y_t(k)$ for the proxy average over the audited rounds of $k$, the corrected estimate is
\begin{equation}
    \hat p_t(k)+\hat\beta_t(k)=\hat r_t(k)+\big(\hat p_t(k)-\bar Y_t(k)\big):
    \label{eq:cv}
\end{equation}
the target average is corrected by the gap between the proxy average over all $N_{t-1}(k)$ pulls and over the $M_{t-1}(k)$ audited ones, which is a control-variate estimator. Write $M=M_{t-1}(k)$, $N=N_{t-1}(k)$, and $C=\mathrm{Cov}(Y,Z)$, so that $\sigma_\beta^2=\sigma_p^2+\sigma_r^2-2C$ when $\sigma_p^2$ and $\sigma_r^2$ are the noise variances. The variance of~\eqref{eq:cv} is then
\begin{equation*}
    \frac{\sigma_\beta^2}{M}+\frac{2C-\sigma_p^2}{N}
    =\frac{\sigma_r^2}{M}-\big(2C-\sigma_p^2\big)\Big(\frac1M-\frac1N\Big),
\end{equation*}
so pairing improves on $\hat r_t(k)$, whose variance is $\sigma_r^2/M$, as soon as $C>\sigma_p^2/2$. Estimating the mismatch as $\hat r_t(k)-\hat p_t(k)$ instead would make the corrected estimate equal to $\hat r_t(k)$ and discard the proxy. Let $c^p_t(k)$, $c^\beta_t(k)$, and $c^r_t(k)$ be confidence radii for the three estimators. The first shrinks with $N_{t-1}(k)$, while the other two shrink with $M_{t-1}(k)$. Their form is given in Section~\ref{sec:method-ucbv}. We build a corrected interval and a target interval, which uses the audited target rewards only,
\begin{align}
    I^{\mathrm{corr}}_t(k)&=\hat p_t(k)+\hat\beta_t(k)\pm\big(c^p_t(k)+c^\beta_t(k)\big),\\
    I^{r}_t(k)&=\hat r_t(k)\pm c^r_t(k),
\end{align}
and intersect them through their endpoints,
\begin{align}
    L_t(k)&=\max\{L^{\mathrm{corr}}_t(k),L^{r}_t(k)\},\\
    U_t(k)&=\min\{U^{\mathrm{corr}}_t(k),U^{r}_t(k)\}.
\end{align}
This hard clipping supplements the correction by using direct target evidence as a safeguard when the corrected interval is loose. After a warm-up in which every intervention is played and audited $m_0$ times, the controller plays
\begin{equation}
    A_t\in\arg\max_{k\in[K]}U_t(k).
\end{equation}
 
\subsection{Confidence Radii}
\label{sec:method-ucbv}

Under Assumption~\ref{ass:model}, for a confidence level $1-\delta$, with $\delta\in(0,1)$, let $\ell_{T,\delta}=\log(6KT/\delta)$. We use
\begin{align}
    c^p_t(k)&=\sigma_p\sqrt{\frac{2\ell_{T,\delta}}{N_{t-1}(k)}},\\
    c^\beta_t(k)&=\sigma_\beta\sqrt{\frac{2\ell_{T,\delta}}{M_{t-1}(k)}},\qquad
    c^r_t(k)=\sigma_r\sqrt{\frac{2\ell_{T,\delta}}{M_{t-1}(k)}}.
\end{align}
The scales $\sigma_p,\sigma_\beta,\sigma_r$ are upper bounds on the conditional sub-Gaussian parameters, not empirical standard deviations.
 
\subsection{Query Gate}
\label{sec:method-gate}
 
An audit is useful only if it can change the decision. Let $J_t\in\arg\max_{j\neq A_t}U_t(j)$ be the most optimistic competitor of the selected intervention. The query gate is
\begin{equation}
    G_t=\mathbf{1}\{L_t(A_t)\le U_t(J_t)\}\;\mathbf{1}\{c^\beta_t(A_t)\ge\rho_q\,c^p_t(A_t)\},
    \label{eq:gate}
\end{equation}
with threshold $\rho_q>0$. The first factor is a certification test: if it is zero, the lower bound of the selected intervention already exceeds the upper bound of every competitor, and no target observation is needed. The second factor compares the two sources of uncertainty. An audited pull reduces $c^\beta_t(A_t)$, whereas a non-audited pull only reduces $c^p_t(A_t)$. The gate therefore requests target feedback only when mismatch uncertainty is at least $\rho_q$ times proxy uncertainty. The gate is computed after the choice of $A_t$ and before the current feedback is observed. Section~\ref{sec:method-pacing} describes how $G_t$ is turned into an audit decision $Q_t$ under the budget $B$.

\subsection{Audit Pacing}
\label{sec:method-pacing}

The warm-up acquires $T_0=Km_0$ audits, and we assume $B\ge T_0$. Let $B_{t-1}=\sum_{s<t}Q_s$ be the number of audits acquired before round $t$, $b_t=B-B_{t-1}$ the remaining budget, and $h_t=T-t+1$ the remaining horizon. The unpaced policy, MC, audits every query-eligible round while the budget lasts:
\begin{equation}
    Q_t=G_t\,\mathbf{1}\{b_t>0\}.
    \label{eq:mc}
\end{equation}
The gate decides \emph{which} decisions deserve an audit, but not \emph{when} the budget is spent. If the relation between proxy and target is stationary, an early audit is as informative as a late one, and auditing as soon as the gate fires is natural. During training, however, this relation may change, and MC exhausts $B$ whenever uncertain decisions are frequent early in training: every later eligible round then goes unaudited, and a later change cannot be detected (Proposition~\ref{prop:frontload}). MC-Paced accepts an eligible audit with probability equal to the remaining budget per remaining round:
\begin{equation}
    Q_t=G_t\,\mathbf{1}\{V_t\le\pi_t\},\quad
    \pi_t=\mathbf{1}\{b_t>0\}\min\Big\{1,\frac{b_t}{h_t}\Big\},
    \label{eq:paced}
\end{equation}
where $(V_t)_t$ are i.i.d.\ uniform on $[0,1]$ and independent of everything else. When an audit is deferred, the selected intervention is still played, and its proxy reward is used to update the controller's estimates. Pacing spreads the available target feedback over the horizon. MC and MC-Paced use the cumulative statistics of Section~\ref{sec:method-index}. Algorithm~\ref{alg:mc-paced} summarizes MC-Paced.

\begin{algorithm}[tb]
    \caption{MC-Paced}
    \label{alg:mc-paced}
    \textbf{Input}: interventions $[K]$, horizon $T$, audit budget $B\ge Km_0$\\
    \textbf{Parameter}: warm-up size $m_0\ge 1$, gate threshold $\rho_q>0$
    \begin{algorithmic}[1]
        \STATE Play and audit every intervention $m_0$ times.
        \FOR{$t=Km_0+1,\dots,T$}
        \STATE Compute $\hat p_t,\hat\beta_t,\hat r_t$ (over $W_t$ for SW-MC-Paced), the radii, and the clipped bounds $L_t(k),U_t(k)$.
        \STATE Play $A_t\in\arg\max_k U_t(k)$ and draw $\mathcal{B}_t\sim w_{A_t}$.
        \STATE Compute $G_t$ (Eq.~\ref{eq:gate}) and $\pi_t$ (Eq.~\ref{eq:paced}), then set $Q_t=G_t\mathbf{1}\{V_t\le\pi_t\}$.
        \STATE Update both models on $\mathcal{B}_t$ and observe $Y_t$.
        \STATE \textbf{if} $Q_t=1$ \textbf{then} observe $Z_t$ and $D_t=Z_t-Y_t$.
        \STATE Update the statistics of $A_t$.
        \ENDFOR
    \end{algorithmic}
\end{algorithm}

\subsection{Sliding-Window Statistics}
\label{sec:method-window}

When the relation between proxy and target may change, SW-MC-Paced runs MC-Paced on statistics computed over the last $\tau$ rounds only. For the window $W_t=\{s:t-\tau\le s\le t-1\}$, let $N^\tau_{t-1}(k)$ and $M^\tau_{t-1}(k)$ be the numbers of pulls and audits of $k$ in $W_t$, and let $\hat p^\tau_t(k)$, $\hat\beta^\tau_t(k)$, and $\hat r^\tau_t(k)$ be the averages of Section~\ref{sec:method-index} restricted to $W_t$. The radii use the same formulas with the window counts and $\ell_\tau=\log(6KT\tau/\delta)$. A radius is $+\infty$ when its count is zero, which encourages exploration of arms without recent observations and allows the gate to request new audits. The intervals, the gate, and the pacing rule are unchanged.

%======================================================================
\section{Analysis}
\label{sec:analysis}

We first analyze MC and MC-Paced when the relation between proxy and target is stationary (Section~\ref{sec:analysis-stationary}), then a sliding-window version when this relation changes during training (Section~\ref{sec:analysis-piecewise}). For $k\neq k^\star$, let
\begin{equation}
    H^p_k=\frac{8(1+\rho_q)^2\sigma_p^2\,\ell}{\Delta_k^2},\qquad
    H^\beta_k=\frac{8(1+\rho_q)^2\sigma_\beta^2\,\ell}{\rho_q^2\Delta_k^2},
\end{equation}
with $\ell=\ell_{T,\delta}$ and $\Delta_{\max}=\max_k\Delta_k$. $H^p_k$ and $H^\beta_k$ are the observation thresholds beyond which proxy uncertainty and mismatch uncertainty can no longer hide the gap $\Delta_k$. They play the same role as the $\log T/\Delta_k^2$ sample thresholds in UCB. All proofs are in Appendix~\ref{app:proofs}. The concentration tools are standard. The contribution is the analysis of the gate, the pacing rule, and their interaction with forgetting.

\subsection{Stationary Case}
\label{sec:analysis-stationary}

Let $\mathcal{E}_T$ be the event on which, for every $t\le T$ and every arm with $M_{t-1}(k)\ge1$, each estimator lies within its radius of its mean: $|\hat p_t(k)-p_k|\le c^p_t(k)$, $|\hat\beta_t(k)-\beta_k|\le c^\beta_t(k)$, and $|\hat r_t(k)-r_k|\le c^r_t(k)$.

\begin{proposition}[Validity]
\label{prop:validity}
$\mathbb{P}(\mathcal{E}_T)\ge1-\delta$. On $\mathcal{E}_T$, $r_k\in I^{\mathrm{corr}}_t(k)\cap I^r_t(k)$ for every such $t$ and $k$, and an arm $a$ with $L_t(a)>\max_{j\neq a}U_t(j)$ is necessarily $k^\star$.
\end{proposition}

A post-warm-up round is \emph{query-eligible} if $G_t=1$, and \emph{deferred} if moreover $Q_t=0$. Let $\Phi_T(k)=\sum_{t>T_0}\mathbf{1}\{A_t=k\}G_t(1-Q_t)$ count the deferred pulls of arm $k$.

\begin{theorem}[Regret decomposition]
\label{thm:decomposition}
Let $m_0\ge1$ and $B\ge T_0$. For MC and MC-Paced, on $\mathcal{E}_T$, every $k\neq k^\star$ satisfies $N_T(k)\le m_0+2+H^p_k+H^\beta_k+\Phi_T(k)$. Hence
\begin{equation}
    R_T\le\sum_{k\neq k^\star}\Delta_k\big(m_0+2+H^p_k+H^\beta_k\big)+\sum_{k\neq k^\star}\Delta_k\Phi_T(k).
\end{equation}
\end{theorem}

The first sum bounds the regret incurred while reducing proxy and mismatch uncertainty. The second bounds the additional regret from suboptimal pulls for which the gate requested an audit but no audit was performed. If every eligible round is audited, $\Phi_T(k)=0$ and, with $\delta=T^{-2}$, $\mathbb{E}[R_T]=O(\log T)$ for fixed gaps. The coefficient $(1+\rho_q)^2(\sigma_p^2+\sigma_\beta^2/\rho_q^2)$ is then minimized by $\rho_q=(\sigma_\beta/\sigma_p)^{2/3}$ (Corollary~\ref{cor:unpaced}, Appendix~\ref{app:proofs}). Under stationarity, the decomposition therefore favors spending audits as soon as the gate fires: an early audit is as informative as a late one, and a deferral can only add regret. This is what MC does. The budget must then cover the requests of the gate, which, for a fixed threshold, may grow linearly with $T$ (Proposition~\ref{prop:cap}, Appendix~\ref{app:proofs}). Under a hard cap, pacing restores a logarithmic bound for any budget proportional to the horizon, at the price of a factor $2/q$ on the mismatch term (Theorem~\ref{thm:paced-app}). Under stationarity, pacing is thus a safeguard for the budget rather than a source of information. The next two results bound the number of audits that any policy needs.

\begin{proposition}[Proxy-only allocation]
\label{prop:proxy}
Let the proxy-optimal arm $k^\star_p=\arg\max_kp_k$ be unique and differ from $k^\star$. Any proxy-only policy satisfies
\[
\mathbb{E}[R_T]\ge\Delta_{k^\star_p}\Big(T-\sum_{k\neq k^\star_p}\mathbb{E}[N_T(k)]\Big).
\]
In particular, proxy-UCB has $\mathbb{E}[R_T]\ge\Delta_{k^\star_p}T-O(\log T)$.
\end{proposition}

Proxy-only allocation fails because it learns the proxy \emph{well}: the better it identifies $k^\star_p$, the closer its target regret is to linear. The bound does not depend on the amount of proxy feedback, so as soon as the two rankings differ, target information is necessary. The next result quantifies how much.

\begin{proposition}[Lower bound on the number of audits]
\label{prop:insufficient}
There exist two-arm instances $\nu_+,\nu_-$ satisfying Assumption~\ref{ass:model}, with identical proxy feedback and target gap $\Delta$, such that any policy using at most $b$ audits in expectation under $\nu_+$ satisfies $\max_\pm\mathbb{E}_\pm[R_T]\ge\frac{\Delta T}{2}\big[1-\frac{\Delta\sqrt b}{2\sigma_\beta}\big]_+$. In particular, a policy with $\max_\pm\mathbb{E}_\pm[R_T]<\Delta T/4$ must use more than $\sigma_\beta^2/\Delta^2$ audits in expectation.
\end{proposition}

In these instances the proxy is uninformative, so only audits can tell which arm is better, and a policy with fewer than $\sigma_\beta^2/\Delta^2$ audits errs on a constant fraction of the rounds. The requirement matches the audit usage of MC: on $\mathcal{E}_T$, each suboptimal arm is audited at most $H^\beta_k+1=O(\sigma_\beta^2\log T/\Delta_k^2)$ times after the warm-up (Appendix~\ref{app:proofs}), the same order up to the logarithmic factor. Propositions~\ref{prop:proxy} and~\ref{prop:insufficient} thus delimit the problem: proxy feedback alone cannot fix a rank-inverting mismatch, and the number of audits it takes to fix it cannot be made arbitrarily small.

\subsection{Piecewise-Stationary Case}
\label{sec:analysis-piecewise}

During training, the relation between the two models may change, and an audit spent early may say nothing about a later regime. We model this by piecewise-stationary means.

\begin{assumption}[Piecewise-stationary paired feedback]
\label{ass:piecewise}
There are $\Gamma_T$ breakpoints $1<b_1<\dots<b_{\Gamma_T}\le T$, fixed in advance, such that the means $p_t(k)$ and $r_t(k)$ are constant on each segment between consecutive breakpoints. Within each segment, the noise conditions of Assumption~\ref{ass:model} hold with the means of that segment. The target-optimal arm $k^\star_t$ is unique at every round, and $\Delta=\min_t\min_{k\neq k^\star_t}\big(r_t(k^\star_t)-r_t(k)\big)>0$.
\end{assumption}

The regret is measured against the best arm of the current segment, $R_T=\sum_{t=1}^T\big(r_t(k^\star_t)-r_t(A_t)\big)$, and $\Delta_{\max}$ now denotes the largest gap over rounds and arms. Under change, spending audits as soon as possible is no longer safe.

\begin{proposition}[Front-loaded audits cannot follow a late change]
\label{prop:frontload}
Fix $\tau_0\in\{2,\dots,T\}$. There exist two environments satisfying Assumption~\ref{ass:piecewise}, with at most one breakpoint, that coincide before $\tau_0$, whose mismatch inverts the target ranking at $\tau_0$ in one of them, and such that any policy using, in the environment without change, at most $b$ audits in expectation on rounds $t\ge\tau_0$ satisfies $\max_\pm\mathbb{E}_\pm\big[\sum_{t\ge\tau_0}(r_t(k^\star_t)-r_t(A_t))\big]\ge\frac{\Delta(T-\tau_0+1)}{2}\big[1-\frac{\Delta\sqrt b}{2\sigma_\beta}\big]_+$.
\end{proposition}

A policy that exhausts its budget before $\tau_0$, as MC does when early decisions are uncertain, has $b=0$ and incurs regret at least $\Delta(T-\tau_0+1)/2$ on one of the two environments. This holds for any rule that decides which rounds to audit, since the change is invisible in the proxy feedback. Pacing keeps audits available after any change, whatever the gate does. Let $B'=B-T_0$, $T'=T-T_0$, and $q=B'/T'\in(0,1]$ be the post-warm-up audit rate.

\begin{lemma}[Pacing preserves the budget]
\label{lem:budget}
For MC-Paced with any gate, including that of SW-MC-Paced, and for every $t>T_0$, $\mathbb{P}(b_t<q h_t/2)\le\exp(-q h_t/8)$. Consequently, $\mathbb{E}\big[\#\{t>T_0:\pi_t<q/2\}\big]\le 8/q$.
\end{lemma}

The proof couples MC-Paced with the policy whose gate always fires. For that policy, the rule $\pi_t=b_t/h_t$ is selection sampling: the audited rounds form a uniformly random subset of size $B'$, so the remaining budget is hypergeometric and concentrates around $qh_t$. A gate that fires less often only leaves more budget. Sliding-window statistics (Section~\ref{sec:method-window}) discard observations from earlier regimes once they fall outside the window. Define $H^p$ and $H^\beta$ as above with $\Delta_k$ replaced by $\Delta$ and $\ell$ by $\ell_{\tau}$.

\begin{theorem}[SW-MC-Paced under piecewise stationarity]
\label{thm:sw}
Under Assumption~\ref{ass:piecewise}, let $m_0\ge1$ and $T_0<B\le T$. SW-MC-Paced with window $\tau$ satisfies
\begin{multline}
    \mathbb{E}[R_T]\le\Delta_{\max}\Big(T_0+\Gamma_T\tau+\frac8q+\delta T\\
    +K\Big\lceil\frac{T}{\tau}\Big\rceil\Big(H^p+1+\frac2q\big(H^\beta+1\big)\Big)\Big).
\end{multline}
\end{theorem}

The structure is that of sliding-window UCB~\cite{garivier2011switching}: $\Gamma_T\tau$ bounds the rounds whose window straddles a change, and each of the $\lceil T/\tau\rceil$ blocks of $\tau$ rounds pays a stationary cost. Pacing enters only through the factor $2/q$ on the mismatch term and the $8/q$ rounds on which the pacing probability is small. With $\tau=T$ and $\Gamma_T=0$, the window keeps the whole history and the bound becomes logarithmic in $T$. Up to the use of the smallest gap $\Delta$, it recovers the stationary bound of Theorem~\ref{thm:paced-app}, the factor $2/q$ being the price of pacing when no change occurs.

\begin{corollary}
\label{cor:sw}
Let $m_0$ be fixed, $T\ge6K$, $\delta=1/T$, $1\le\Gamma_T\le T$, and $\tau=\lceil\sqrt{T\log T/\Gamma_T}\rceil$. Then $\mathbb{E}[R_T]=O\big(\Delta_{\max}\,C_q\sqrt{\Gamma_TT\log T}\big)$, where
\begin{equation*}
    C_q=1+\frac Kq+\frac{K(1+\rho_q)^2}{\Delta^2}\Big(\sigma_p^2+\frac{\sigma_\beta^2}{q\rho_q^2}\Big).
\end{equation*}
\end{corollary}

For a fixed audit rate $q$, the paced sliding-window policy thus reaches the $\sqrt{\Gamma_TT\log T}$ rate of sliding-window UCB, whereas by Proposition~\ref{prop:frontload} a policy that runs out of audits before a change has linear regret. Pacing is what makes the guarantee possible: Lemma~\ref{lem:budget} keeps audits available in every window, and the window keeps the corresponding estimates current.

%======================================================================
\section{Experiments}
\label{sec:experiments}

The experiments run MC and MC-Paced with cumulative statistics, together with discounted variants, rather than the sliding-window policy of Section~\ref{sec:analysis-piecewise}. They test the regimes discussed in the analysis, including a mismatch that drifts during training, in synthetic environments and in real training runs.

\subsection{Synthetic Environments: A Map of Regimes}
\label{sec:exp-synthetic}

\begin{figure*}[t]
    \centering
    \includegraphics[width=\textwidth]{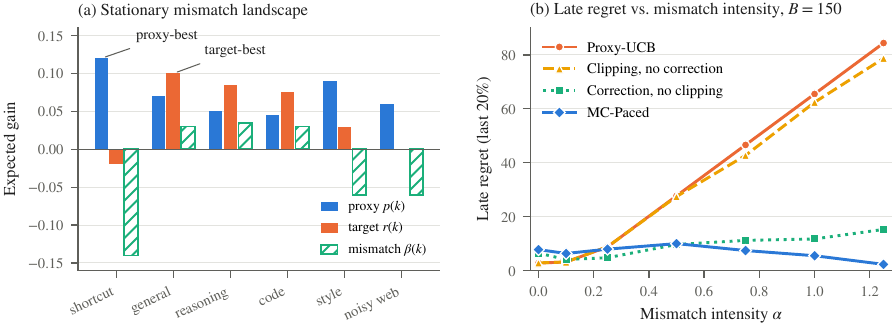}
    \caption{Synthetic environments. (a) Stationary mismatch landscape: the proxy prefers \textit{shortcut}, the target prefers \textit{general}, and the mismatch $\beta(k)=r(k)-p(k)$ depends on the arm and inverts the ranking. (b) Late target regret (last $20\%$ of the horizon) as a function of the mismatch intensity $\alpha$, with $B=150$. For small $\alpha$ the proxy already ranks the arms well and correction only adds calibration cost. Once the mismatch changes the ranking, Proxy-UCB degrades linearly while the variants that use the paired residual remain low.}
    \label{fig:synthetic}
\end{figure*}

\paragraph{Setup.}
We first test the mechanism where proxy and target utilities are known for every arm. The environments use $K=6$ synthetic groups and a horizon $T=3000$. Playing arm $k$ applies the mixture $w(k)=(1-\lambda)e_k+\lambda\pi_0$, where $e_k$ puts all probability mass on group $k$ and $\pi_0$ is a reference distribution. The proxy reward is $Y_t=p_t(A_t)+\xi_t$ and, on audited rounds, the target reward is $Z_t=r_t(A_t)+\zeta_t$, where $\xi_t$ and $\zeta_t$ denote the proxy and target observation noise, respectively. Since $p_t$ and $r_t$ are controlled, the target regret is computed exactly. Under non-stationarity we also report the late regret over the last $20\%$ of the horizon, $R_{\mathrm{late}}=\sum_{t=\lfloor0.8T\rfloor+1}^T\big(\max_kr_t(k)-r_t(A_t)\big)$, which isolates the quality of decisions once initialization effects have faded. Three regimes are considered: \emph{aligned} ($p_t=r_t$, hence $\beta_t=0$), \emph{stationary mismatch}, in which the proxy-best and target-best arms differ (Figure~\ref{fig:synthetic}a), and \emph{smooth mismatch}, in which $p_t$ and $r_t$ drift over the horizon so that $\beta_t$ is itself non-stationary. The proxy and target noise levels are $\sigma_p=0.03$ and $\sigma_r=0.05$, the mixture uses $\lambda=0.2$ and a uniform $\pi_0$, and every configuration is run on 30 seeds. The main budget is $B=150$. Proxy-UCB denotes a UCB policy that uses proxy rewards only.

\paragraph{When the proxy is aligned, a proxy-only policy suffices.}
In the stationary aligned control, Proxy-UCB is nearly optimal, with a late regret of $0.03$, while policies that spend target audits pay for their calibration. In the smooth aligned control, Proxy-UCB and MC-Paced end with similar late regrets ($9.89$ and $9.91$), but MC-Paced keeps a higher total regret ($51.06$ against $23.01$). Correction is not free: its value depends on the existence of a mismatch.

\paragraph{A rank-inverting mismatch makes the proxy fail.}
In the stationary mismatch regime, the proxy prefers \textit{shortcut} while the target prefers \textit{general} (Figure~\ref{fig:synthetic}a). Proxy-UCB accumulates a late regret of $71.95$, as predicted by Proposition~\ref{prop:proxy}. Methods that use the paired residual reduce it by an order of magnitude: MC reaches $3.10$, and the ablation that keeps the correction without clipping reaches $3.15$. Because $\beta$ does not drift, an early calibration suffices: a target-only baseline that spends its audits at the beginning even reaches $1.90$. Pacing becomes useful when early audits no longer represent the current proxy--target relation.

\paragraph{Under drift, correction must stay up to date.}
In the smooth mismatch regime with $B=150$, Proxy-UCB fails ($R_{\mathrm{late}}=65.51$). MC spends its audits too early and still reaches $28.67$, whereas MC-Paced keeps audits until the late rounds and reaches $5.50$. A policy that spreads target audits without using the correction reaches $14.16$, and clipping without correction stays close to Proxy-UCB ($62.43$). The two comparisons isolate two effects: the residual correction is needed to repair the ranking, and pacing becomes decisive when that correction can become stale. The schedule explains the difference: with the same budget, a front-loaded schedule consumes all audits around $t\approx149$, whereas the paced one keeps audits until the end of the horizon. Proposition~\ref{prop:frontload} formalizes why a schedule that runs out of audits cannot follow a late change.

\paragraph{Varying the mismatch intensity.}
To check that these results do not depend on one favorable environment, we interpolate the target between the proxy and a base target, $r_\alpha(t,k)=p_t(k)+\alpha\big(r_{\mathrm{base}}(t,k)-p_t(k)\big)$, so that $\alpha=0$ means perfect alignment (Figure~\ref{fig:synthetic}b). At $\alpha=0$, Proxy-UCB obtains a late regret of $2.77$ against $7.77$ for MC-Paced. The order reverses as the mismatch grows: $27.70$ against $10.00$ at $\alpha=0.5$, $65.51$ against $5.50$ at $\alpha=1$, and $84.42$ against $2.31$ at $\alpha=1.25$. The ablations give the same reading: removing the correction while keeping clipping does not fix the failure, whereas removing clipping while keeping the correction remains competitive, especially for moderate mismatch. Clipping is therefore a conservative safeguard rather than the source of the gain.

\paragraph{Varying the audit budget.}
In the smooth mismatch regime with $B\in\{6,12,30,60,150,300\}$, the calibration error of MC-Paced decreases with the budget, from a mean absolute error of $0.0504$ at $B=6$ to $0.0186$ at $B=150$, and its late regret from $22.54$ to $5.50$. At very small budgets the relation is not monotone: a few audits can move the ranking without yet stabilizing it. Between $B=150$ and $B=300$ the late regret stays comparable ($5.50$ against $6.08$) while the total regret keeps decreasing, so the marginal value of extra audits for late decisions becomes small in this environment.

\subsection{CivilComments-WILDS Setup}
\label{sec:exp-setup}
 
\paragraph{Task and data.}
We instantiate the framework on Civil\-Comments-WILDS, a binary toxicity classification benchmark \cite{borkan2019nuanced,koh2021wilds} with identity annotations designed to study subpopulation shift. This choice does not reproduce the adaptation of a generative LLM. It provides a controllable closed loop in which data groups are interpretable, worst-group robustness is measurable, and two models of different capacity are actually trained on the same blocks. After cleaning, the training pool contains 266{,}224 examples.
 
\paragraph{Interventions.}
The action space, denoted B8, contains eight interventions corresponding to the identities \textit{male}, \textit{female}, \textit{LGBTQ}, \textit{christian}, \textit{muslim}, \textit{other religions}, \textit{black}, and \textit{white}. These groups overlap. Each block contains 512 examples: 256 drawn from the selected group, label-balanced (128 non-toxic and 128 toxic), and 256 drawn from the natural distribution of the pool. A block corresponds to 16 target updates.
 
\paragraph{Proxy and target.}
The target model is BERT-Large (24 layers, hidden size 1024, about 335M parameters)~\cite{devlin2019bert}. The proxy is BERT-Small (4 layers, hidden size 512, about 28.8M parameters)~\cite{turc2019wellread}, selected by a preliminary scan over five BERT sizes (Appendix~\ref{app:details}). The proxy reward is the decrease in class-balanced cross-entropy of BERT-Small on a fixed proxy validation set $V_{\mathrm{proxy}}$ of 2{,}001 examples, measured every round. The target reward is the same quantity for BERT-Large on the full target validation set $V_{\mathrm{target}}$ of 20{,}005 examples, with $V_{\mathrm{proxy}}\subset V_{\mathrm{target}}$, measured only on audited rounds. The mismatch therefore combines a model gap and an evaluation-set gap.
 
\paragraph{Metrics and information contract.}
The main metric is the worst-group accuracy (WGA) over the sixteen cells formed by the eight identities crossed with the two labels, denoted WGA-16~\cite{sagawa2020distributionally}. Macro cross-entropy over the 16 identity-label cells (denoted CE in the tables) and overall accuracy are reported as guardrails. The controller never observes WGA: its target reward is a local cross-entropy gain, whereas WGA-16 measures the robustness of the final model. Trajectory metrics are computed on an independent validation set $V_{\mathrm{val}}$ of 24{,}780 examples that never updates the controller. The official test set (133{,}782 examples) stays closed during development and is evaluated only on frozen checkpoints. All policies within a seed start from the same checkpoint, and methods that use target feedback share the same audit cap. Unlike the synthetic setting, the exact regret is not observable here, since counterfactual rewards of unplayed interventions are unknown. Policies are therefore compared through the state actually reached by the target model.

\subsection{Correction at Equal Target Budget}
\label{sec:exp-correction}
 
We first evaluate short late-stage adaptation. Both models start from a common checkpoint trained for about two epoch-equivalents, and the controlled phase adds only 0.12 to 0.25 epoch-equivalent. In these campaigns the corrected policy is MC-DUCB, which keeps the mismatch-corrected index but uses discounted statistics, as in discounted UCB~\cite{garivier2011switching}, and audits on a schedule fixed in advance instead of using the gate and pacing: after the first audit, each new audit is placed at the round that is farthest from the audits already scheduled, so that target feedback covers the whole phase (Appendix~\ref{app:details}). MC-DUCB is not covered by the analysis of Section~\ref{sec:analysis}: these campaigns test the correction principle, while Section~\ref{sec:exp-timing} tests the gate and the pacing rule. It is compared with Static, a balanced round-robin over the eight interventions that receives no target feedback, Proxy-only, which decides from proxy rewards only, and Sparse-Target, which decides from the sparse target rewards alone and audits on the same schedule as MC-DUCB. In these campaigns, all policies use statistics discounted with a half-life of 64 rounds ($\gamma=2^{-1/64}$).
 
\paragraph{Ten paired seeds at equal budget.}
H64 denotes a trajectory of 64 rounds, including 40 shared burn-in rounds, and target-aware policies may reveal $B=48$ target rewards. During the burn-in, each of the eight interventions is played $m_0=5$ times in a balanced order. The budget counts every target reward revealed over the trajectory, burn-in included. H128 uses 128 rounds, including the same 40-round burn-in. For instance, $B=128$ reveals the target reward at every round. Table~\ref{tab:h64} reports the final metrics. At identical target information, MC-DUCB improves WGA-16 over Sparse-Target by $+2.52$ points (95\% bootstrap CI $[+0.96,+4.28]$, exact paired sign-flip test $p=0.018$, $p_{\mathrm{Holm}}=0.035$ over the three main contrasts) and over Proxy-only by $+2.51$ points (positive on nine of ten seeds, $p_{\mathrm{Holm}}=0.023$), without a resolved difference in accuracy. The contrast with Static ($+1.57$ points, $p_{\mathrm{Holm}}=0.061$) is suggestive only. The gain therefore does not come from more target information but from a different use of the same audits: the dense proxy provides temporal coverage and the audits correct its preferences. Consistently, the per-seed correlation between the mean residual of an intervention and its induced reallocation relative to Proxy-only is positive on all ten seeds, showing that the correction changes which interventions are played.
 
\begin{table}[t]
    \centering
    \small
    \setlength{\tabcolsep}{3pt}
    \begin{tabular}{lrrr}
        \toprule
        Method & CE $\downarrow$ & WGA-16 $\uparrow$ & Acc. $\uparrow$ \\
        \midrule
        Proxy-only            & .4493\,{\scriptsize$\pm$.0064} & .6564\,{\scriptsize$\pm$.0232} & .9111\,{\scriptsize$\pm$.0032} \\
        Sparse-Target, $B=48$ & .4500\,{\scriptsize$\pm$.0076} & .6563\,{\scriptsize$\pm$.0322} & .9099\,{\scriptsize$\pm$.0044} \\
        Static                & .4452\,{\scriptsize$\pm$.0027} & .6658\,{\scriptsize$\pm$.0257} & .9098\,{\scriptsize$\pm$.0027} \\
        MC-DUCB, $B=48$       & .4452\,{\scriptsize$\pm$.0038} & \textbf{.6815}\,{\scriptsize$\pm$.0197} & .9095\,{\scriptsize$\pm$.0024} \\
        \bottomrule
    \end{tabular}
    \caption{Short adaptation (H64), mean $\pm$ standard deviation over ten paired seeds. Sparse-Target and MC-DUCB observe the same number of target rewards.}
    \label{tab:h64}
\end{table}
 
\paragraph{Budget sweep.}
The exploratory H128 campaign (three seeds, 88 adaptive rounds) varies the audit budget. At the three budgets shared with Sparse-Target, MC-DUCB obtains a higher WGA-16 ($+2.96$, $+4.99$, and $+1.51$ points at $B=16$, $32$, and $64$, see Table~\ref{tab:h128}), and a lower cross-entropy at $B=16$ and $32$. The best endpoint of the sweep is MC-DUCB at $B=32$ (WGA-16 of .6985), above the dense configuration, MC-DUCB with $B=128$, which reveals the target reward at every round (.6905), and the relation between budget and final WGA is not monotone. With three seeds, we read this campaign as evidence that a moderate number of paired audits suffices, not as the identification of an optimal budget.
 
\begin{table}[t]
    \centering
    \small
    \begin{tabular}{rrr}
        \toprule
        $B$ & $\Delta$WGA-16 (pts) & $\Delta$CE \\
        \midrule
        16 & $+2.96$ & $-0.0033$ \\
        32 & $+4.99$ & $-0.0070$ \\
        64 & $+1.51$ & $+0.0019$ \\
        \bottomrule
    \end{tabular}
    \caption{H128: MC-DUCB minus Sparse-Target at equal target budget, mean over three seeds.}
    \label{tab:h128}
\end{table}

\subsection{Audit Timing under Long Training}
\label{sec:exp-timing}
 
\paragraph{Protocol.}
We next run long trajectories in which the controller takes over from natural-distribution training at different stages. An epoch-equivalent is 520 blocks (8{,}320 target updates). All regimes end after two epoch-equivalents. They differ only by the natural prefix: the controller starts from the pretrained model (n0), after 0.25 epoch-equivalent (n25), or after one epoch-equivalent (n100), which gives 1{,}040, 910, and 520 controlled rounds. The audit cap is about 25\% of the controlled rounds (260, 228, and 130 audits). Besides MC and MC-Paced, the long runs include Static, Proxy-DUCB (a discounted UCB on proxy rewards only), Sparse-Target, MC-DUCB, and MC-Paced-Forget, which is MC-Paced with discounted statistics as in discounted UCB~\cite{garivier2011switching} (half-life of 64 rounds, $\gamma=2^{-1/64}$) and serves as an ablation of memory. Results are averaged over five paired seeds.
 
\paragraph{The proxy is informative but imperfect.}
On the audited rounds of MC-Paced, proxy and target rewards have a Pearson correlation of about 0.52, a Spearman correlation of about 0.45, and a sign agreement of about 66\% in all three regimes. The proxy thus carries real high-frequency information, but about one third of the gains have opposite signs for the two models, which is the intermediate regime targeted by our formulation.
 
\paragraph{Pacing at identical audit count.}
We isolate the temporal component by comparing MC and MC-Paced, which share the proxy, the cumulative residual estimator, hard clipping, and the audit cap, and which end up consuming exactly the same number of audits. Table~\ref{tab:pacing} shows that pacing improves test WGA-16 by $+6.32$, $+7.20$, and $+10.54$ points in n0, n25, and n100 (95\% bootstrap CIs $[+5.87,+6.94]$, $[+5.90,+8.67]$, $[+9.31,+11.77]$), with gains on all five seeds in every regime. The independent validation set shows the same pattern ($+7.76$, $+7.67$, $+12.60$ points). With five seeds, the evidence rests on effect size, seed-level consistency, and replication across the two evaluation sets rather than on an exact sign test.
 
\begin{table}[t]
    \centering
    \small
    \begin{tabular}{llrrr}
        \toprule
        Regime & Method & CE $\downarrow$ & WGA-16 $\uparrow$ & Acc. $\uparrow$ \\
        \midrule
        n0   & MC        & .918 & .579 & .924 \\
             & MC-Paced  & .901 & \textbf{.642} & .916 \\
        n25  & MC        & .928 & .589 & .923 \\
             & MC-Paced  & .829 & \textbf{.661} & .915 \\
        n100 & MC        & .732 & .624 & .921 \\
             & MC-Paced  & .552 & \textbf{.729} & .913 \\
        \bottomrule
    \end{tabular}
    \caption{Official test set, mean over five paired seeds. MC and MC-Paced consume the same number of audits in each regime.}
    \label{tab:pacing}
\end{table}
 
\begin{figure}[t]
    \centering
    \includegraphics[width=\columnwidth]{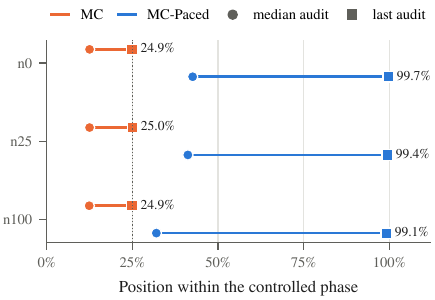}
    \caption{Position of the audits within the controlled phase on CivilComments-WILDS. Circles mark the median audit and squares the last audit. Both methods consume the same cap. MC spends it in the first quarter of the phase, whereas MC-Paced keeps audits until the end.}
    \label{fig:audit-timing}
\end{figure}

\paragraph{Mechanism.}
Without pacing, MC meets enough uncertain decisions early to exhaust its cap at about 25\% of the controlled phase, with a median audit at about 12.5\%. MC-Paced spends the same number of audits but places its median audit at 42.6\%, 41.2\%, and 32.0\% of the phase and its last audit at 99--100\% (Figure~\ref{fig:audit-timing}). The allocation follows: the normalized entropy of the realized allocation is 0.39, 0.38, and 0.34 for MC, against 0.96, 0.96, and 0.94 for MC-Paced. A consistent interpretation is that a calibration built from early, still uncertain audits locks MC onto a few interventions, and that this decision cannot be revised once the budget is spent. Pacing reserves part of the target feedback budget for later stages, when the proxy--target relation may have drifted.
 
\paragraph{Where the gain comes from.}
Relative to MC, pacing raises the test accuracy of the toxic cells ($y=1$) by $+5.96$, $+6.62$, and $+7.98$ points while lowering that of the non-toxic cells by 3.7 to 4.5 points, at a cost of about 0.8 point of overall accuracy. The method thus trades part of the performance on already well-handled cells for robustness on the hardest ones. Adding exponential forgetting (half-life 64 rounds) on top of pacing does not help: it ends 0.17, 1.24, and 0.70 points of test WGA-16 below MC-Paced. In this benchmark, the robust component is the timing of the audits rather than an additional memory parameter.

\subsection{Comparison with Baselines}
\label{sec:exp-baselines}

Table~\ref{tab:full-test} (Appendix~\ref{app:results}) reports every policy on the official test set, and Table~\ref{tab:contrasts} the paired contrasts of MC-Paced against each baseline. The comparison reveals three regimes rather than a single dominant method.

\paragraph{n0: the proxy is hard to beat.}
When control starts from the pretrained model, the best controlled policies remain close in WGA-16: MC-DUCB obtains the best mean ($.646$), ahead of MC-Paced ($.642$), Sparse-Target ($.640$), and Proxy-DUCB ($.638$). MC-Paced has the lowest cross-entropy among them ($.901$). A dense proxy already provides a strong decision signal in this regime.

\paragraph{n25: pacing improves the corrected policy.}
After a short natural prefix, MC-Paced obtains both the best WGA-16 ($.661$) and the lowest cross-entropy ($.829$). Against MC-DUCB, the fixed-schedule corrected policy, it gains $+2.63$ points of WGA-16 (95\% CI $[+0.68,+5.08]$) and lowers cross-entropy by $0.058$ (CI $[-0.108,-0.017]$), with a positive WGA gain on four of five seeds.

\paragraph{n100: a shared robustness front.}
When control starts after one natural epoch-equivalent, Proxy-DUCB, MC-Paced, and MC-DUCB reach the same robustness front: the paired gap between MC-Paced and Proxy-DUCB is $-0.15$ point (CI $[-1.11,+0.55]$). MC-Paced keeps a lower cross-entropy than Static, Sparse-Target, and MC-DUCB, and gains $+0.71$ point over Sparse-Target (CI $[+0.11,+1.31]$). Every controlled policy except MC improves WGA-16 by $18$ to $19$ test points over the starting checkpoint ($.541$). This gain comes from switching to a controlled allocation, not from a particular controller.

Overall, correction and pacing make MC-Paced competitive in all three regimes while avoiding the collapse of the unpaced policy. This is consistent with the role assigned to pacing by the analysis.

%======================================================================
\section{Limitations and Conclusion}
\label{sec:conclusion}

\paragraph{Limitations.}
The proof of concept uses a pair of BERT models (Small and Large) on a classification benchmark rather than generative language models, eight interventions, and between three and ten seeds per campaign. The controller optimizes a local cross-entropy gain while we report worst-group accuracy, which measures a different property of the final model. The audit budget limits target information, not compute: the target model is trained on every block. The empirical comparison uses a single benchmark and baselines that isolate the components of the method (fixed allocation, proxy-only, target-only, and fixed-schedule correction). Comparisons with data allocation methods such as DoReMi, DoGE, or ODM~\cite{xie2023doremi,fan2024doge,albalak2023odm}, and with Group DRO~\cite{sagawa2020distributionally} for worst-group robustness, are left to an extended version. On the theoretical side, the guarantee under change covers abrupt, piecewise-stationary changes and a sliding-window policy whose window is tuned with the number of changes. The experiments run its cumulative and discounted counterparts, MC-Paced and MC-Paced-Forget, and smooth drift is covered only empirically. All upper bounds assume known noise scales.

\paragraph{Conclusion.}
A small model can guide the data allocation of a larger one while ranking the data differently. Paired audits make this arm-dependent mismatch identifiable, and a mismatch-corrected bandit uses them to repair the proxy ranking. When the relation between the two models is stationary, audits are best spent as soon as the gate requests them. When this relation changes, a policy that has already exhausted its audit budget cannot detect a later change from proxy feedback alone. Pacing combined with a sliding window reaches the rate of sliding-window UCB. The experiments show that, at identical audit count, the timing of audits matters as much as their number. Natural next steps are an evaluation of the sliding-window policy, windows that adapt to an unknown number of changes, variance-adaptive confidence radii that do not require known noise scales, and experiments on language model data mixing.

%======================================================================
\appendix
 
\section{Proofs}
\label{app:proofs}

Throughout, $\mathcal{F}^{\mathrm{pre}}_t$ denotes the sigma-field generated by $\mathcal{H}_{t-1}$, the selected arm $A_t$, and every quantity used to compute $G_t$ and $\pi_t$, but not $V_t$ nor the current feedback. Under Assumption~\ref{ass:model} or~\ref{ass:piecewise} and Eq.~\eqref{eq:paced}, $\mathbb{E}[Q_t\mid\mathcal{F}^{\mathrm{pre}}_t]=G_t\pi_t$ for MC-Paced and SW-MC-Paced.

\subsection{Proof of Proposition~\ref{prop:validity}}
Fix an arm $k$ and let $\tau_{k,i}$ be the round of its $i$-th pull. Since $A_t$ is predictable, $(Y_{\tau_{k,i}}-p_k)_i$ is a martingale-difference sequence with conditionally $\sigma_p$-sub-Gaussian increments, so for every deterministic $n\ge1$,
\[
\mathbb{P}\Big(\Big|\frac1n\sum_{i=1}^n(Y_{\tau_{k,i}}-p_k)\Big|>\sigma_p\sqrt{2\ell/n}\Big)\le2e^{-\ell}.
\]
Because $Q_t$ is chosen before the current feedback is observed, the paired residual noises and the target noises of the audited pulls of arm $k$ form martingale-difference sequences with $\sigma_\beta$- and $\sigma_r$-sub-Gaussian increments, and satisfy the same bound. A union bound over at most $T$ sample sizes, three estimators, and $K$ arms gives total failure probability $3KT\cdot2e^{-\ell}=\delta$, hence $\mathbb{P}(\mathcal{E}_T)\ge1-\delta$. On $\mathcal{E}_T$, $|\hat p_t(k)+\hat\beta_t(k)-r_k|\le c^p_t(k)+c^\beta_t(k)$ and $|\hat r_t(k)-r_k|\le c^r_t(k)$, so $r_k$ lies in both intervals and $L_t(k)\le r_k\le U_t(k)$. If $L_t(a)>\max_{j\neq a}U_t(j)$, then $r_a\ge L_t(a)>U_t(j)\ge r_j$ for every $j\neq a$, hence $a=k^\star$. \qed

\subsection{Proof of Theorem~\ref{thm:decomposition}}
Work on $\mathcal{E}_T$ and consider a post-warm-up round with $A_t=k\neq k^\star$. Optimism, the action rule, and the fact that clipping can only lower the corrected upper bound give $r_{k^\star}\le U_t(k^\star)\le U_t(k)\le U^{\mathrm{corr}}_t(k)\le r_k+2c^p_t(k)+2c^\beta_t(k)$, that is,
\begin{equation}
    \Delta_k\le2\big(c^p_t(k)+c^\beta_t(k)\big).
    \label{eq:gapradii}
\end{equation}
By Proposition~\ref{prop:validity}, the selected arm is not certified, so the first factor of $G_t$ equals one.

\emph{Proxy-dominant pulls} ($c^\beta_t(k)<\rho_qc^p_t(k)$). Then \eqref{eq:gapradii} gives $\Delta_k<2(1+\rho_q)\sigma_p\sqrt{2\ell/N_{t-1}(k)}$, i.e.\ $N_{t-1}(k)<H^p_k$. Each such pull increases $N_t(k)$, so there are at most $H^p_k+1$ of them.

\emph{Mismatch-dominant pulls} ($c^\beta_t(k)\ge\rho_qc^p_t(k)$). Then $G_t=1$, and \eqref{eq:gapradii} gives $\Delta_k\le2(1+1/\rho_q)\sigma_\beta\sqrt{2\ell/M_{t-1}(k)}$, i.e.
\begin{equation}
    M_{t-1}(k)\le H^\beta_k.
    \label{eq:eligiblecount}
\end{equation}
This holds whether or not the round is audited. Each audited pull increases $M_t(k)$, so there are at most $H^\beta_k+1$ audited mismatch-dominant pulls. The others have $G_t=1$ and $Q_t=0$ and are counted by $\Phi_T(k)$. Adding the $m_0$ warm-up pulls gives $N_T(k)\le m_0+2+H^p_k+H^\beta_k+\Phi_T(k)$, and $R_T=\sum_{k\neq k^\star}\Delta_kN_T(k)$ concludes. \qed

\begin{corollary}[Fully served gate]
\label{cor:unpaced}
If every query-eligible round is audited, then $\Phi_T(k)=0$ for all $k$, and with $\delta=T^{-2}$, $\mathbb{E}[R_T]=O\big(\sum_{k\neq k^\star}(1+\rho_q)^2(\sigma_p^2+\sigma_\beta^2/\rho_q^2)\log(KT)/\Delta_k\big)$. The coefficient is minimized at $\rho_q=(\sigma_\beta/\sigma_p)^{2/3}$.
\end{corollary}

\begin{proof}
$G_t=1$ implies $Q_t=1$, so $\Phi_T(k)=0$ and Theorem~\ref{thm:decomposition} applies on $\mathcal{E}_T$. Off $\mathcal{E}_T$, $R_T\le\Delta_{\max}T$, which contributes at most $\Delta_{\max}/T$ in expectation. For $f(\rho)=(1+\rho)^2(a+b/\rho^2)$ with $a,b>0$, $f'(\rho)=2(1+\rho)(a-b/\rho^3)$, whose unique positive root $\rho=(b/a)^{1/3}$ is the global minimizer since $f\to\infty$ at $0$ and $\infty$. Take $a=\sigma_p^2$, $b=\sigma_\beta^2$.
\end{proof}

\subsection{Proof of Proposition~\ref{prop:proxy}}
Every pull of $k^\star_p$ costs $\Delta_{k^\star_p}>0$, so
\[
R_T\ge\Delta_{k^\star_p}N_T(k^\star_p)=\Delta_{k^\star_p}\Big(T-\sum_{k\neq k^\star_p}N_T(k)\Big).
\] Taking expectations gives the first claim. The standard fixed-gap bound $\sum_{k\neq k^\star_p}\mathbb{E}[N_T(k)]=O(\log T)$ for UCB on the proxy means~\cite{auer2002finite} gives the second. \qed

\subsection{Proof of Proposition~\ref{prop:insufficient}}
Take two arms and let every proxy observation be $Y_t\sim\mathcal{N}(0,\sigma_p^2)$ independently of the arm. An audited pull also reveals $D_t$, conditionally independent of $Y_t$, with $Z_t=Y_t+D_t$. Under $\nu_+$, $D_t\sim\mathcal{N}(\Delta,\sigma_\beta^2)$ for arm $1$ and $\mathcal{N}(0,\sigma_\beta^2)$ for arm $2$. Under $\nu_-$ the two means are swapped. Both instances satisfy Assumption~\ref{ass:model} with $p_k=0$ and $\sigma_r^2=\sigma_p^2+\sigma_\beta^2$. The proxy channel contributes no divergence, and each audit contributes $\Delta^2/(2\sigma_\beta^2)$. By the chain rule, $\mathrm{KL}(\mathbb{P}_+\|\mathbb{P}_-)\le\Delta^2b/(2\sigma_\beta^2)$, and Pinsker's inequality gives $\mathrm{TV}(\mathbb{P}_+,\mathbb{P}_-)\le\Delta\sqrt b/(2\sigma_\beta)$. For every round, $\mathbb{P}_+(A_t=2)+\mathbb{P}_-(A_t=1)\ge1-\mathrm{TV}$. Summing over $t$, $\mathbb{E}_+[R_T]+\mathbb{E}_-[R_T]\ge\Delta T(1-\mathrm{TV})$, and the maximum is at least half the sum. \qed

\subsection{Proof of Lemma~\ref{lem:budget}}
Index post-warm-up rounds by $i=t-T_0\in\{1,\dots,T'\}$, so that $h_t=T'-i+1$. The warm-up consumes exactly $T_0$ audits, hence $b_{T_0+1}=B'$. Write $\pi(b,h)=\mathbf{1}\{b>0\}\min\{1,b/h\}$, which is nondecreasing in $b$.

\emph{Coupling.} Define the reference process $\bar b_{T_0+1}=B'$, $\bar Q_t=\mathbf{1}\{V_t\le\pi(\bar b_t,h_t)\}$, $\bar b_{t+1}=\bar b_t-\bar Q_t$, driven by the same uniforms as MC-Paced. We show $b_t\ge\bar b_t$ by induction. If $b_t\ge\bar b_t+1$, then $b_{t+1}\ge b_t-1\ge\bar b_t\ge\bar b_{t+1}$. If $b_t=\bar b_t$, then $\pi_t=\pi(\bar b_t,h_t)$ and $Q_t=G_t\mathbf{1}\{V_t\le\pi_t\}\le\bar Q_t$, so $b_{t+1}\ge\bar b_{t+1}$.

\emph{The reference process is selection sampling.} Since $B'\le T'$, induction gives $\bar b_t\le h_t$ for all $t$ (if $\bar b_t=h_t$, the round is selected with probability one), so the selected set $S\subseteq\{1,\dots,T'\}$ always has size $B'$. For a fixed $S$ with $|S|=B'$, the probability of selecting exactly $S$ is a product of factors $\bar b/h$ over $i\in S$ and $(h-\bar b)/h$ over $i\notin S$. Along this path the numerators over $S$ run through $B',B'-1,\dots,1$, those over the complement through $T'-B',\dots,1$, and the denominators through $T',\dots,1$. The probability is therefore $B'!(T'-B')!/T'!=\binom{T'}{B'}^{-1}$: $S$ is a uniformly random subset of size $B'$. Consequently $\bar b_t=|S\cap\{i,\dots,T'\}|$ is hypergeometric: the number of marked items among $h_t$ draws without replacement from $T'$ items of which $B'$ are marked, with mean $qh_t$.

\emph{Concentration.} By Hoeffding's comparison theorem for sampling without replacement~\cite{hoeffding1963probability}, $\mathbb{E}[e^{-\lambda\bar b_t}]\le\mathbb{E}[e^{-\lambda X}]$ for every $\lambda\ge0$, with $X\sim\mathrm{Bin}(h_t,q)$. The multiplicative Chernoff bound for the binomial lower tail therefore applies: $\mathbb{P}(\bar b_t\le qh_t/2)\le\exp(-qh_t/8)$. Since $b_t\ge\bar b_t$, $\mathbb{P}(b_t<qh_t/2)\le\exp(-qh_t/8)$.

\emph{Counting.} If $b_t\ge qh_t/2>0$, then $\pi_t\ge q/2$, since $\pi_t=\min\{1,b_t/h_t\}$. Hence
$\mathbb{E}[\#\{t>T_0:\pi_t<q/2\}]\le\sum_{h=1}^{T'}e^{-qh/8}\le\frac{1}{e^{q/8}-1}\le\frac8q$. \qed

\subsection{Audit Requests and Pacing under Stationarity}

The following results justify the discussion after Theorem~\ref{thm:decomposition}.

\begin{proposition}[The gate does not bound the number of audits]
\label{prop:cap}
For every arm,
\[
[M_T(k)-m_0]_+\le1+\frac{\sigma_\beta^2N_T(k)}{\rho_q^2\sigma_p^2};
\]
hence
\[
B_T\le\min\Big\{B,\,K(m_0+1)+\frac{\sigma_\beta^2T}{\rho_q^2\sigma_p^2}\Big\}.
\]
\end{proposition}

\begin{proof}
If arm $k$ receives no post-warm-up audit, the claim is immediate. Otherwise let $\tau_k$ be its last post-warm-up audited round. Since $Q_{\tau_k}=1$, the gate was open, so $c^\beta_{\tau_k}(k)\ge\rho_qc^p_{\tau_k}(k)$. Substituting the radii gives $M_{\tau_k-1}(k)\le\sigma_\beta^2N_{\tau_k-1}(k)/(\rho_q^2\sigma_p^2)$. Hence $M_T(k)=M_{\tau_k-1}(k)+1\le1+\sigma_\beta^2N_T(k)/(\rho_q^2\sigma_p^2)$. Summing over arms with $\sum_kN_T(k)=T$ and adding the warm-up audits gives the bound on $B_T$. The hard cap gives $B_T\le B$.
\end{proof}

For a fixed threshold, the gate may thus request a number of audits linear in $T$, and Corollary~\ref{cor:unpaced} does not apply to MC once its budget is exhausted. Pacing removes this dependence on the budget.

\begin{theorem}[MC-Paced under stationarity]
\label{thm:paced-app}
Under Assumption~\ref{ass:model}, let $m_0\ge1$ and $T_0<B\le T$. MC-Paced satisfies
\begin{multline*}
    \mathbb{E}[R_T]\le(m_0+2)\sum_{k\neq k^\star}\Delta_k+\frac{8\Delta_{\max}}{q}+\delta\Delta_{\max}T\\
    +8(1+\rho_q)^2\ell\sum_{k\neq k^\star}\frac{\sigma_p^2+2\sigma_\beta^2/(q\rho_q^2)}{\Delta_k}.
\end{multline*}
In particular, with $\delta=T^{-2}$, fixed gaps, and $B'\ge cT'$ for some $c\in(0,1]$, $\mathbb{E}[R_T]=O(\log(T)/c)$.
\end{theorem}

\begin{proof}
For $k\neq k^\star$ and $t>T_0$, let $e_t(k)=\mathbf{1}\{A_t=k,\,G_t=1,\,M_{t-1}(k)\le H^\beta_k\}$, which is $\mathcal{F}^{\mathrm{pre}}_t$-measurable, and $\tilde\Phi_T(k)=\sum_{t>T_0}e_t(k)(1-Q_t)$.

\emph{Step 1 (reduction).} By \eqref{eq:eligiblecount}, every deferred pull of $k$ on $\mathcal{E}_T$ has $M_{t-1}(k)\le H^\beta_k$, so $\Phi_T(k)=\tilde\Phi_T(k)$ on $\mathcal{E}_T$. Off $\mathcal{E}_T$, $R_T\le\Delta_{\max}T$. Theorem~\ref{thm:decomposition} therefore gives
\begin{multline*}
\mathbb{E}[R_T]\le\sum_{k\neq k^\star}\Delta_k\big(m_0+2+H^p_k+H^\beta_k\big)\\
+\sum_{k\neq k^\star}\Delta_k\mathbb{E}[\tilde\Phi_T(k)]+\delta\Delta_{\max}T.
\end{multline*}

\emph{Step 2 (audits available while the mismatch is unresolved).} Each round with $e_t(k)Q_t=1$ increases $M_t(k)$ by one and requires $M_{t-1}(k)\le H^\beta_k$, starting from $M_{T_0}(k)=m_0$. Hence, pathwise, $\sum_te_t(k)Q_t\le s_k:=(\lfloor H^\beta_k\rfloor-m_0+1)_+\le H^\beta_k$, where the last inequality uses $m_0\ge1$. Since $\mathbb{E}[Q_t\mid\mathcal{F}^{\mathrm{pre}}_t]=\pi_t$ when $e_t(k)=1$, $\mathbb{E}\big[\sum_te_t(k)\pi_t\big]\le H^\beta_k$.

\emph{Step 3 (deferrals).} Similarly,
\[
\mathbb{E}[\tilde\Phi_T(k)]=\mathbb{E}\Big[\sum_te_t(k)(1-\pi_t)\Big].
\]
If $\pi_t\ge q/2$, then $1-\pi_t\le(2/q-1)\pi_t$. Therefore
\[
\mathbb{E}[\tilde\Phi_T(k)]\le\Big(\frac2q-1\Big)H^\beta_k+\mathbb{E}\Big[\sum_te_t(k)\mathbf{1}\{\pi_t<q/2\}\Big].
\]

\emph{Step 4 (rounds with a small pacing probability).} At most one arm is played per round, so $\sum_{k\neq k^\star}\Delta_k\sum_te_t(k)\mathbf{1}\{\pi_t<q/2\}\le\Delta_{\max}\#\{t>T_0:\pi_t<q/2\}$, whose expectation is at most $8\Delta_{\max}/q$ by Lemma~\ref{lem:budget}.

\emph{Step 5.} Combining, the mismatch contribution of arm $k$ is $\Delta_kH^\beta_k\big(1+2/q-1\big)=8(1+\rho_q)^2\ell\cdot2\sigma_\beta^2/(q\rho_q^2\Delta_k)$, and $\Delta_kH^p_k=8(1+\rho_q)^2\ell\sigma_p^2/\Delta_k$, which gives the statement. With $\delta=T^{-2}$, $\ell=O(\log T)$ and $\delta\Delta_{\max}T=\Delta_{\max}/T$. The condition $q\ge c$ gives the last claim.
\end{proof}

\subsection{Proof of Proposition~\ref{prop:frontload}}
Let both environments follow the instance $\nu_+$ of the proof of Proposition~\ref{prop:insufficient} on rounds $t<\tau_0$. From round $\tau_0$ on, the first environment keeps following $\nu_+$, while the second follows $\nu_-$: its mismatch, hence its target-optimal arm, changes at $\tau_0$. Before $\tau_0$, the two environments have identical observation kernels and contribute no divergence. From $\tau_0$ on, the proxy channel is identical and each audit contributes $\Delta^2/(2\sigma_\beta^2)$. The chain rule and Pinsker's inequality give $\mathrm{TV}\le\Delta\sqrt b/(2\sigma_\beta)$, where $b$ bounds the expected number of audits on rounds $t\ge\tau_0$ in the first environment. For every $t\ge\tau_0$, the probabilities of playing the post-change suboptimal arm in the two environments sum to at least $1-\mathrm{TV}$. Summing over $t\ge\tau_0$ and taking the larger of the two regrets proves the claim. \qed

\subsection{Proof of Theorem~\ref{thm:sw}}

\emph{Good rounds.} Call a round $t$ \emph{good} if no breakpoint lies in $\{t-\tau+1,\dots,t\}$, so that the means are constant on $W_t\cup\{t\}$. Each breakpoint makes at most $\tau$ rounds bad, so there are at most $\Gamma_T\tau$ bad rounds.

\emph{Concentration.} Fix an arm and one of the three estimators. Under Assumption~\ref{ass:piecewise}, the noises of the successive pulls (or audits) of this arm form a martingale-difference sequence with conditionally sub-Gaussian increments. By optional skipping, the average of the noises over any fixed range of $n$ consecutive pull (or audit) indices exceeds $\sigma\sqrt{2\ell_\tau/n}$ in absolute value with probability at most $2e^{-\ell_\tau}$. A union bound over $K$ arms, three estimators, at most $T$ starting indices, and at most $\tau$ lengths gives probability at most $6KT\tau e^{-\ell_\tau}=\delta$. On the complementary event $\mathcal{E}^\tau_T$, at every good round the window averages are within their radii of the current means, because the pulls of $W_t$ form a range of consecutive pull indices of the arm and all lie in the current segment.

\emph{Suboptimal pulls on good rounds.} Work on $\mathcal{E}^\tau_T$ and consider a good post-warm-up round with $A_t=k\neq k^\star_t$. The argument of Theorem~\ref{thm:decomposition}, with infinite radii satisfying every inequality trivially, gives $\Delta\le r_t(k^\star_t)-r_t(k)\le2\big(c^p_t(k)+c^\beta_t(k)\big)$ with window counts, and the selected arm is not certified. If $c^\beta_t(k)<\rho_qc^p_t(k)$, then $N^\tau_{t-1}(k)<H^p$. Otherwise $G_t=1$ and $M^\tau_{t-1}(k)\le H^\beta$.

\emph{Counting in blocks.} Split $\{1,\dots,T\}$ into $\lceil T/\tau\rceil$ blocks of $\tau$ consecutive rounds. Within a block, every earlier round lies in the window of every later one. Hence, if arm $k$ is pulled at $j$ rounds of a block, each with $N^\tau_{t-1}(k)<H^p$, the last of them has $N^\tau_{t-1}(k)\ge j-1$, so $j\le H^p+1$. Likewise, at most $H^\beta+1$ audited pulls of $k$ per block have $M^\tau_{t-1}(k)\le H^\beta$. Over all blocks, proxy-dominant pulls of $k$ number at most $\lceil T/\tau\rceil(H^p+1)$, and audited mismatch-dominant pulls at most $\lceil T/\tau\rceil(H^\beta+1)$.

\emph{Deferred pulls.} Let $e_t(k)=\mathbf{1}\{t\text{ good},\,t>T_0,\,A_t=k\neq k^\star_t,\,G_t=1,\,M^\tau_{t-1}(k)\le H^\beta\}$, which is $\mathcal{F}^{\mathrm{pre}}_t$-measurable. As in Steps~2--4 of the proof of Theorem~\ref{thm:paced-app}, $\mathbb{E}[\sum_te_t(k)\pi_t]=\mathbb{E}[\sum_te_t(k)Q_t]\le\lceil T/\tau\rceil(H^\beta+1)$, the inequality $1-\pi_t\le(2/q-1)\pi_t$ holds when $\pi_t\ge q/2$, and Lemma~\ref{lem:budget}, which holds for any gate, bounds the rounds with $\pi_t<q/2$ by $8/q$ in expectation over all arms. The expected number of deferred mismatch-dominant pulls, summed over arms, is therefore at most $K\lceil T/\tau\rceil(2/q-1)(H^\beta+1)+8/q$.

\emph{Conclusion.} Each round costs at most $\Delta_{\max}$. Adding the $T_0$ warm-up rounds, the $\Gamma_T\tau$ bad rounds, at most $T$ rounds on the complement of $\mathcal{E}^\tau_T$, whose probability is at most $\delta$, and the three counts above, whose mismatch terms sum to $(1+2/q-1)(H^\beta+1)=(2/q)(H^\beta+1)$ per arm and block, gives the statement. \qed

\subsection{Proof of Corollary~\ref{cor:sw}}
Since $\Gamma_T\le T$, we have $\tau\le T$. With $\delta=1/T$ and $T\ge6K$, $\ell_\tau=\log(6KT^2\tau)\le4\log T$ and $\delta T=1$, so
\begin{align*}
    H^p&\le\frac{32(1+\rho_q)^2\sigma_p^2\log T}{\Delta^2},\\
    H^\beta&\le\frac{32(1+\rho_q)^2\sigma_\beta^2\log T}{\rho_q^2\Delta^2}.
\end{align*}
With $\tau=\lceil\sqrt{T\log T/\Gamma_T}\rceil$, we have $\Gamma_T\tau\le\sqrt{\Gamma_TT\log T}+\Gamma_T\le2\sqrt{\Gamma_TT\log T}$ and $\lceil T/\tau\rceil\le\sqrt{\Gamma_TT/\log T}+1$. Hence both $\lceil T/\tau\rceil$ and $\lceil T/\tau\rceil\log T$ are $O(\sqrt{\Gamma_TT\log T})$. In Theorem~\ref{thm:sw}, the terms $T_0+\Gamma_T\tau+\delta T$ are therefore $O(\sqrt{\Gamma_TT\log T})$. The terms $K\lceil T/\tau\rceil(1+2/q)$ and $8/q$ are $O\big((K/q)\sqrt{\Gamma_TT\log T}\big)$. The terms in $H^p$ and $H^\beta$ are $O\big(K(1+\rho_q)^2(\sigma_p^2+\sigma_\beta^2/(q\rho_q^2))\Delta^{-2}\sqrt{\Gamma_TT\log T}\big)$. \qed

\section{Experimental Details and Hyperparameters}
\label{app:details}
 
\paragraph{Proxy selection.}
Before any closed-loop experiment, we compared five BERT sizes as candidate proxies for BERT-Large by measuring how faithfully each ranks the interventions (Table~\ref{tab:proxy-scan}). The last column is a one-shot decision gap, $\Delta_{\mathrm{top1}}=\max_k r(k)-r(\arg\max_j p(j))$, not a cumulative regret. BERT-Small evaluated on 10\% of the target validation set was chosen as an intermediate regime: much lighter than the target and informative enough to guide allocation (top-1 agreement 1.00, Spearman 0.92), yet imperfect on local variations (28.3\% sign error). This scan is performed before training and does not measure the mismatch encountered along the adaptive trajectory.
 
\begin{table}[h]
    \centering
    \small
    \setlength{\tabcolsep}{4pt}
    \begin{tabular}{lrrrr}
        \toprule
        Proxy & Top-1 & Spearman & Sign err. & $\Delta_{\mathrm{top1}}$ \\
        \midrule
        Tiny (4.4M)            & 0.667 & 0.933 & 0.000 & 0.0126 \\
        Mini (11.2M)           & 0.667 & 0.933 & 0.500 & 0.0021 \\
        Small (28.8M) + 10\%   & 1.000 & 0.920 & 0.283 & 0.0000 \\
        Medium (41.4M)         & 1.000 & 0.867 & 0.250 & 0.0000 \\
        Base (109.5M)          & 1.000 & 0.867 & 0.167 & 0.0000 \\
        \bottomrule
    \end{tabular}
    \caption{Preliminary fidelity scan of candidate proxies against BERT-Large (335.1M).}
    \label{tab:proxy-scan}
\end{table}
 
\paragraph{Frozen protocol.}
Table~\ref{tab:config} lists the frozen parameters of the CivilComments-WILDS protocol.
 
\begin{table*}[t]
    \centering
    \small
    \begin{tabular}{p{0.22\textwidth}p{0.72\textwidth}}
        \toprule
        Component & Configuration \\
        \midrule
        Training pool & 266{,}224 examples after cleaning, with frozen indices. \\
        Proxy validation & 2{,}001 fixed examples ($\approx$10\% of the target validation set), evaluated with BERT-Small every round. \\
        Target validation & 20{,}005 examples, evaluated with BERT-Large only when the target reward is revealed. \\
        Independent validation & 24{,}780 examples, used for trajectory metrics and never visible to the controller. \\
        Official test & 133{,}782 examples, opened only after checkpoints were frozen. \\
        Proxy / target models & BERT-Small (4 layers, 512 hidden, 28.8M) / BERT-Large (24 layers, 1024 hidden, 335M). \\
        Action space B8 & Eight overlapping interventions: male, female, LGBTQ, christian, muslim, other religions, black, white. \\
        Training block & 512 examples: 256 from the selected group (128 with $y=0$, 128 with $y=1$), 256 from the natural distribution. Each block produces 16 target updates. \\
        Pairing & The block chosen by $A_t$ updates both models. The budget limits revealed target rewards, not target updates. \\
        Rewards & Decrease of class-balanced cross-entropy on $V_{\mathrm{proxy}}$ (proxy) and $V_{\mathrm{target}}$ (target). The residual is $D_t=Z_t-Y_t$. \\
        Main metric & WGA-16 over identity $\times$ label cells, with macro-16 CE and accuracy as guardrails. \\
        Epoch-equivalent & 520 blocks = 8{,}320 target updates. Common endpoint E2 = 1{,}040 blocks. \\
        Long regimes & n0: 1{,}040 controlled rounds from E0, n25: 910 rounds from natural round 130, and n100: 520 rounds from E1. \\
        Audit cap & $\approx$25\% of the controlled phase: 260, 228, 130 audits for n0, n25, n100. \\
        MC-Paced & Cumulative statistics, query gate, pacing $\pi_t=\min\{1,b_t/h_t\}$ with remaining budget $b_t$ and horizon $h_t$. \\
        Forgetting ablation & Exponential forgetting with half-life $H=64$, $\gamma=2^{-1/64}$. \\
        Controller & $\rho_q=1$, with a warm-up of $m_0=5$ rounds per intervention ($T_0=40$), audited in the long runs. $B$ counts all audits. \\
        Fixed schedule & Sparse-Target and MC-DUCB: farthest-point audit rounds (first after warm-up, then last round, then midpoints). \\
        Discount (H64, H128) & All statistics, half-life 64 rounds. \\
        Seeds & 10 (H64), 3 (H128), 5 (long regimes), paired across methods. \\
        \bottomrule
    \end{tabular}
    \caption{Frozen configuration of the CivilComments-WILDS proof of concept.}
    \label{tab:config}
\end{table*}
 
\paragraph{Baselines.}
Static is a balanced round-robin over the eight interventions and receives no target feedback. Proxy-only (H64, H128) and Proxy-DUCB (long runs) decide from proxy rewards only with a discounted UCB rule. Sparse-Target, called Gold-only in the H64 campaign, decides from the sparse target rewards alone. MC-DUCB uses the mismatch-corrected index with discounted statistics. Dense-Target denotes MC-DUCB with a budget equal to the horizon, so that the target reward is revealed at every round. Natural denotes training on the natural distribution of the pool. In the H64 and H128 campaigns, all statistics are discounted with weight $2^{-(t-s)/64}$ for an observation from round $s$.

\paragraph{Fixed audit schedule.}
Sparse-Target and MC-DUCB audit on a schedule fixed before the run by a farthest-point rule. The first post-warm-up audit takes place at the first round after the warm-up and the second at the last round. Each subsequent audit is placed at the round that maximizes the distance to the nearest audit already scheduled, that is, in the middle of the largest remaining gap. For instance, with $T=2000$ and a warm-up of 40 rounds, the first three audits fall at rounds 41, 2000, and about 1020. Target feedback is thus spread over the whole phase, refined by successive halving.

\paragraph{Controller parameters.}
The gate threshold is $\rho_q=1$, the balanced choice of Corollary~\ref{cor:unpaced} when proxy and residual noise scales are comparable. In all CivilComments campaigns, the first $T_0=40$ rounds play each of the $K=8$ interventions $m_0=5$ times. In the long runs, these warm-up rounds are also audited. In H64 and H128, they form the shared burn-in, and target rewards are revealed within the budget $B$, which counts every audit of the trajectory, so that budgets below 40, such as $B=8$ in H128, are possible. The proxy is \texttt{google/bert\_uncased\_L-4\_H-512\_A-8} and the target \texttt{google-bert/bert-large-uncased}.
 
%----------------------------------------------------------------------
\section{Additional Results}
\label{app:results}
 
\paragraph{Exploratory budget sweep.}
For H128, the main endpoints (three seeds) are: Static .6812, Proxy-only .6754, Sparse-Target ($B=32$) .6486, MC-DUCB ($B=32$) .6985, MC-DUCB ($B=96$) .6795, MC-DUCB ($B=128$, dense) .6905 in WGA-16. Cross-entropy is lowest for Static (.4524) and MC-DUCB at $B=96$ (.4530), and highest for Sparse-Target (.4707).
 
\paragraph{Long regimes: all policies.}
Table~\ref{tab:full-test} reports all policies evaluated on the official test set. Natural-start (n25) and Natural E1 (n100) are the shared checkpoints preceding the controlled phases. In n25, the test evaluations of Static, Proxy-DUCB, and Sparse-Target were completed after the protocol was closed, on frozen checkpoints and without any selection or tuning on the test set.
 
\begin{table}[h]
    \centering
    \small
    \setlength{\tabcolsep}{3pt}
    \begin{tabular}{lrrr}
        \toprule
        Method & CE $\downarrow$ & WGA-16 $\uparrow$ & Acc. $\uparrow$ \\
        \midrule
        \multicolumn{4}{l}{\textit{n0 -- control from the pretrained model}} \\
        Natural E2        & .526\,{\scriptsize$\pm$.010} & .570\,{\scriptsize$\pm$.018} & .929\,{\scriptsize$\pm$.001} \\
        Static            & .968\,{\scriptsize$\pm$.031} & .632\,{\scriptsize$\pm$.021} & .916\,{\scriptsize$\pm$.000} \\
        Proxy-DUCB        & .935\,{\scriptsize$\pm$.026} & .638\,{\scriptsize$\pm$.014} & .916\,{\scriptsize$\pm$.000} \\
        Sparse-Target     & .944\,{\scriptsize$\pm$.017} & .640\,{\scriptsize$\pm$.018} & .917\,{\scriptsize$\pm$.002} \\
        MC-DUCB           & .932\,{\scriptsize$\pm$.023} & .646\,{\scriptsize$\pm$.023} & .916\,{\scriptsize$\pm$.001} \\
        MC                & .918\,{\scriptsize$\pm$.037} & .579\,{\scriptsize$\pm$.010} & .924\,{\scriptsize$\pm$.001} \\
        MC-Paced          & .901\,{\scriptsize$\pm$.061} & .642\,{\scriptsize$\pm$.008} & .916\,{\scriptsize$\pm$.001} \\
        MC-Paced-Forget   & .967\,{\scriptsize$\pm$.019} & .640\,{\scriptsize$\pm$.018} & .917\,{\scriptsize$\pm$.002} \\
        \midrule
        \multicolumn{4}{l}{\textit{n25 -- control after 0.25 epoch-equivalent}} \\
        Natural-start     & .603\,{\scriptsize$\pm$.078} & .392\,{\scriptsize$\pm$.100} & .924\,{\scriptsize$\pm$.002} \\
        Static            & .907\,{\scriptsize$\pm$.050} & .645\,{\scriptsize$\pm$.011} & .916\,{\scriptsize$\pm$.001} \\
        Proxy-DUCB        & .856\,{\scriptsize$\pm$.041} & .655\,{\scriptsize$\pm$.011} & .916\,{\scriptsize$\pm$.001} \\
        Sparse-Target     & .892\,{\scriptsize$\pm$.051} & .634\,{\scriptsize$\pm$.030} & .917\,{\scriptsize$\pm$.002} \\
        MC-DUCB           & .887\,{\scriptsize$\pm$.028} & .634\,{\scriptsize$\pm$.023} & .916\,{\scriptsize$\pm$.002} \\
        MC                & .928\,{\scriptsize$\pm$.039} & .589\,{\scriptsize$\pm$.014} & .923\,{\scriptsize$\pm$.001} \\
        MC-Paced          & .829\,{\scriptsize$\pm$.050} & .661\,{\scriptsize$\pm$.015} & .915\,{\scriptsize$\pm$.001} \\
        MC-Paced-Forget   & .893\,{\scriptsize$\pm$.052} & .648\,{\scriptsize$\pm$.013} & .917\,{\scriptsize$\pm$.002} \\
        \midrule
        \multicolumn{4}{l}{\textit{n100 -- control after Natural E1}} \\
        Natural E1        & .522\,{\scriptsize$\pm$.080} & .541\,{\scriptsize$\pm$.099} & .927\,{\scriptsize$\pm$.003} \\
        Static            & .576\,{\scriptsize$\pm$.026} & .722\,{\scriptsize$\pm$.006} & .914\,{\scriptsize$\pm$.001} \\
        Proxy-DUCB        & .550\,{\scriptsize$\pm$.019} & .731\,{\scriptsize$\pm$.004} & .912\,{\scriptsize$\pm$.001} \\
        Sparse-Target     & .578\,{\scriptsize$\pm$.024} & .722\,{\scriptsize$\pm$.015} & .915\,{\scriptsize$\pm$.001} \\
        MC-DUCB           & .573\,{\scriptsize$\pm$.019} & .727\,{\scriptsize$\pm$.014} & .915\,{\scriptsize$\pm$.002} \\
        MC                & .732\,{\scriptsize$\pm$.021} & .624\,{\scriptsize$\pm$.012} & .921\,{\scriptsize$\pm$.001} \\
        MC-Paced          & .552\,{\scriptsize$\pm$.023} & .729\,{\scriptsize$\pm$.009} & .913\,{\scriptsize$\pm$.002} \\
        MC-Paced-Forget   & .572\,{\scriptsize$\pm$.025} & .722\,{\scriptsize$\pm$.007} & .914\,{\scriptsize$\pm$.001} \\
        \bottomrule
    \end{tabular}
    \caption{Official test set, mean $\pm$ standard deviation over five paired seeds.}
    \label{tab:full-test}
\end{table}
 
\paragraph{Paired contrasts against baselines.}
Table~\ref{tab:contrasts} reports seed-level contrasts of MC-Paced against each baseline on the test set. A negative $\Delta$CE is favorable. No adaptive policy clearly dominates in n0, which is consistent with a dense proxy already providing a strong decision signal when control starts from the pretrained model. In n25, MC-Paced improves both CE and WGA-16 over MC-DUCB ($+2.63$ points, 95\% CI $[+0.68,+5.08]$, and $\Delta$CE $-0.058$, CI $[-0.108,-0.017]$). In n100, MC-Paced lies on the same robustness front as Proxy-DUCB ($-0.15$ point, CI $[-1.11,+0.55]$) and improves CE over Static, Sparse-Target, and MC-DUCB. The large increase of WGA-16 from Natural E1 to the end of the controlled phase in n100 ($+18.83$ test points for MC-Paced) is shared by all controlled policies and should be attributed to switching to a controlled allocation, not to a specific controller.
 
\begin{table}[h]
    \centering
    \small
    \begin{tabular}{llrrr}
        \toprule
        Regime & Baseline & $\Delta$CE & $\Delta$WGA & Wins \\
        \midrule
        n0   & Proxy-DUCB    & $-0.034$ & $+0.40$ & 2/5 \\
             & Sparse-Target & $-0.043$ & $+0.20$ & 2/5 \\
             & MC-DUCB       & $-0.032$ & $-0.39$ & 3/5 \\
             & Static        & $-0.068$ & $+0.97$ & 3/5 \\
        \midrule
        n25  & Static        & $-0.078$ & $+1.58$ & 4/5 \\
             & Proxy-DUCB    & $-0.027$ & $+0.53$ & 4/5 \\
             & Sparse-Target & $-0.063$ & $+2.64$ & 4/5 \\
             & MC-DUCB       & $-0.058$ & $+2.63$ & 4/5 \\
        \midrule
        n100 & Proxy-DUCB    & $+0.003$ & $-0.15$ & 3/5 \\
             & Sparse-Target & $-0.026$ & $+0.71$ & 4/5 \\
             & Static        & $-0.024$ & $+0.75$ & 4/5 \\
             & MC-DUCB       & $-0.021$ & $+0.23$ & 2/5 \\
        \bottomrule
    \end{tabular}
    \caption{MC-Paced minus baseline on the test set, seed-level means over five seeds. $\Delta$WGA is reported in points. Wins counts seeds with a positive $\Delta$WGA.}
    \label{tab:contrasts}
\end{table}
 
\paragraph{Per-cell decomposition of the pacing gain.}
Table~\ref{tab:cells} splits the MC-Paced minus MC contrast between non-toxic ($y=0$) and toxic ($y=1$) cells.
 
\begin{table}[h]
    \centering
    \small
    \setlength{\tabcolsep}{4pt}
    \begin{tabular}{llrrrr}
        \toprule
        Split & Regime & $\Delta$Acc$_{0}$ & $\Delta$Acc$_{1}$ & $\Delta$CE$_{0}$ & $\Delta$CE$_{1}$ \\
        \midrule
        Val.  & n0   & $-3.68$ & $+6.60$ & $+0.142$ & $-0.224$ \\
              & n25  & $-4.42$ & $+7.32$ & $+0.140$ & $-0.365$ \\
              & n100 & $-4.65$ & $+8.60$ & $+0.050$ & $-0.441$ \\
        Test  & n0   & $-3.69$ & $+5.96$ & $+0.141$ & $-0.175$ \\
              & n25  & $-4.13$ & $+6.62$ & $+0.137$ & $-0.334$ \\
              & n100 & $-4.47$ & $+7.98$ & $+0.049$ & $-0.408$ \\
        \bottomrule
    \end{tabular}
    \caption{MC-Paced minus MC by label, with accuracies in points. The robustness gain comes mostly from toxic cells.}
    \label{tab:cells}
\end{table}
 
%----------------------------------------------------------------------
\section{Reproducibility Contract}
\label{app:repro}
 
\paragraph{Separation between decision and evaluation.}
At round $t$, the intervention $A_t$ and the audit decision $Q_t$ are chosen before the current feedback is observed. The proxy reward may update the controller at every round, whereas the target reward updates it only when $Q_t=1$. Metrics computed on the independent validation set $V_{\mathrm{val}}$ are never fed back to the controller and serve only to reconstruct trajectories of cross-entropy, WGA-16, and accuracy. Counterfactual scans used to analyze the mechanism are likewise diagnostic only. The official test set remained closed during development and method selection.
 
\paragraph{Paired comparisons and exact resumption.}
Comparisons between policies are paired by seed: within a regime, all methods start from the same state and follow the same data-construction conventions, so that differences isolate the effect of the allocation policy. Long runs are checkpointed so that they can be resumed exactly: model weights, optimizer states, controller state, round counters, and remaining audit budget are restored jointly, since an incorrect resumption could alter both the trajectory of the model and the information available to the controller. All complementary test evaluations were performed on checkpoints frozen before the test set was opened, and no training, tuning, or checkpoint selection was performed after observing them.
 
\section*{Ethical Statement}
CivilComments-WILDS contains toxic comments and identity annotations. We use it only to study worst-group robustness of toxicity classifiers. No new data were collected. Improving the worst identity group comes with a measurable trade-off: relative to MC, MC-Paced lowers the accuracy of non-toxic cells by 3.7 to 4.5 points, which should be weighed before deployment.

\section*{Acknowledgments}
This work was conducted during Amine Saadi's internship at Orange Research, Lannion.

%======================================================================
\bibliographystyle{named}
\bibliography{references}
 
\end{document}